\documentclass[letterpaper]{article} 
\usepackage{aaai25}  
\usepackage{times}  
\usepackage{helvet}  
\usepackage{courier}  
\usepackage[hyphens]{url}  
\usepackage{graphicx} 
\usepackage{natbib}  
\usepackage{caption} 
\usepackage{algorithm}
\usepackage{algorithmic}

\usepackage{amsmath}
\usepackage{graphicx}
\usepackage{float}
\usepackage{subfigure}
\usepackage{amssymb}

\usepackage{subcaption}
\usepackage{bm}
\usepackage{multirow}
\usepackage{multicol}
\usepackage{booktabs}
\usepackage{makecell}
\usepackage{bbding}
\usepackage{caption}

\usepackage[table]{xcolor}
\usepackage{colortbl}

\usepackage{newfloat}
\usepackage{listings}

\usepackage[switch]{lineno}

\DeclareCaptionStyle{ruled}{labelfont=normalfont,labelsep=colon,strut=off} 
\floatstyle{ruled}
\newfloat{listing}{tb}{lst}{}
\floatname{listing}{Listing}
\title{Fit and Prune:  Fast and Training-free  Visual Token Pruning for\\ Multi-modal Large Language Models}
\author{
    Weihao Ye\textsuperscript{\rm 1\rm 2},
    Qiong Wu\textsuperscript{\rm 1\rm 2},
    Wenhao Lin\textsuperscript{\rm 1\rm 2},
    Yiyi Zhou\textsuperscript{\rm 1\rm 2}\thanks{Corresponding Author.}
}
\affiliations{
    \textsuperscript{\rm 1}Key Laboratory of Multimedia Trusted Perception and Efficient Computing,\\
    Ministry of Education of China, Xiamen University, 361005, P.R. China.\\
    
    \textsuperscript{\rm 2}Institute of Artificial Intelligence, Xiamen University, 361005, P.R. China.\\
    
    \{weihaoye, qiong, wenhaolin\}@stu.xmu.edu.cn, zhouyiyi@xmu.edu.cn
    
}

\usepackage{bibentry}

\begin{document}

\maketitle

\begin{abstract}
Recent progress in \emph{Multimodal Large Language Models} (MLLMs) often use large image tokens to compensate the visual shortcoming of MLLMs, which not only exhibits obvious redundancy but also greatly exacerbates the already high computation. Token pruning is an effective solution for speeding up MLLMs, but  when and how to drop tokens still remains a challenge.  In this paper, we propose a novel and training-free approach for the effective visual token pruning of MLLMs, termed \emph{FitPrune}, which can quickly produce a complete pruning recipe for MLLMs according to a pre-defined budget.  Specifically, FitPrune considers token pruning as a statistical problem of MLLM and its objective is to find out an optimal pruning scheme that can minimize the divergence of the attention distributions before and after pruning. In practice, FitPrune can be quickly accomplished based on the attention statistics from a small batch of inference data, avoiding the expensive trials of MLLMs.
According to the pruning recipe, an MLLM can directly remove the redundant visual tokens of different examples during inference.  To validate FitPrune, we apply it to a set of recent MLLMs, including LLaVA-1.5, LLaVA-HR and LLaVA-NEXT, and conduct extensive experiments on a set of benchmarks. The experimental results show that our FitPrune can not only reduce the computational complexity to a large extent, while retaining high performance, \emph{e.g.}, -54.9\% FLOPs for LLaVA-NEXT with only 0.5\% accuracy drop. Notably, the pruning recipe can be obtained in about 5 minutes. Our code is available at \url{https://github.com/ywh187/FitPrune}.
\end{abstract}

\begin{figure}[!th]
\centering
\includegraphics[width=1.0\linewidth]{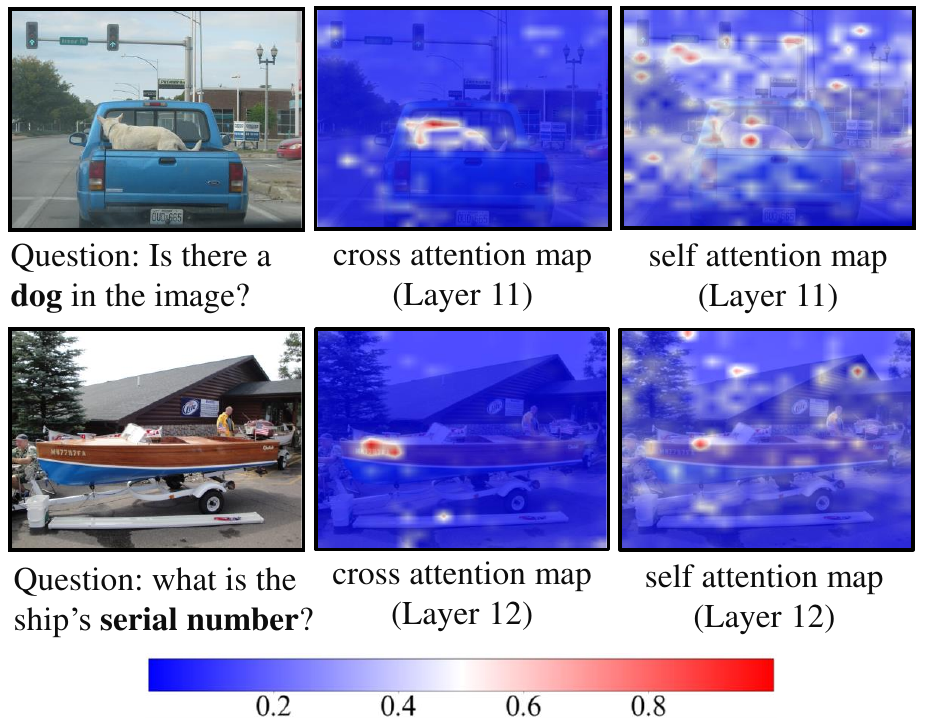}
\vspace{-0.5cm}
\caption{Visualization of  the  cross and self attention of image tokens of LLaVA-1.5 7B~\cite{liu2023visual}. These tokens become less active at the higher layer of an MLLM.}

\vspace{-4mm}
\label{fig:moti_1}
\end{figure}


%



\begin{figure*}[!t]
\centering  
\subfigure[Cross-Attention Weight Distribution]{
\label{Fig.moti2_cross_distribution}
\includegraphics[width=0.491\textwidth]{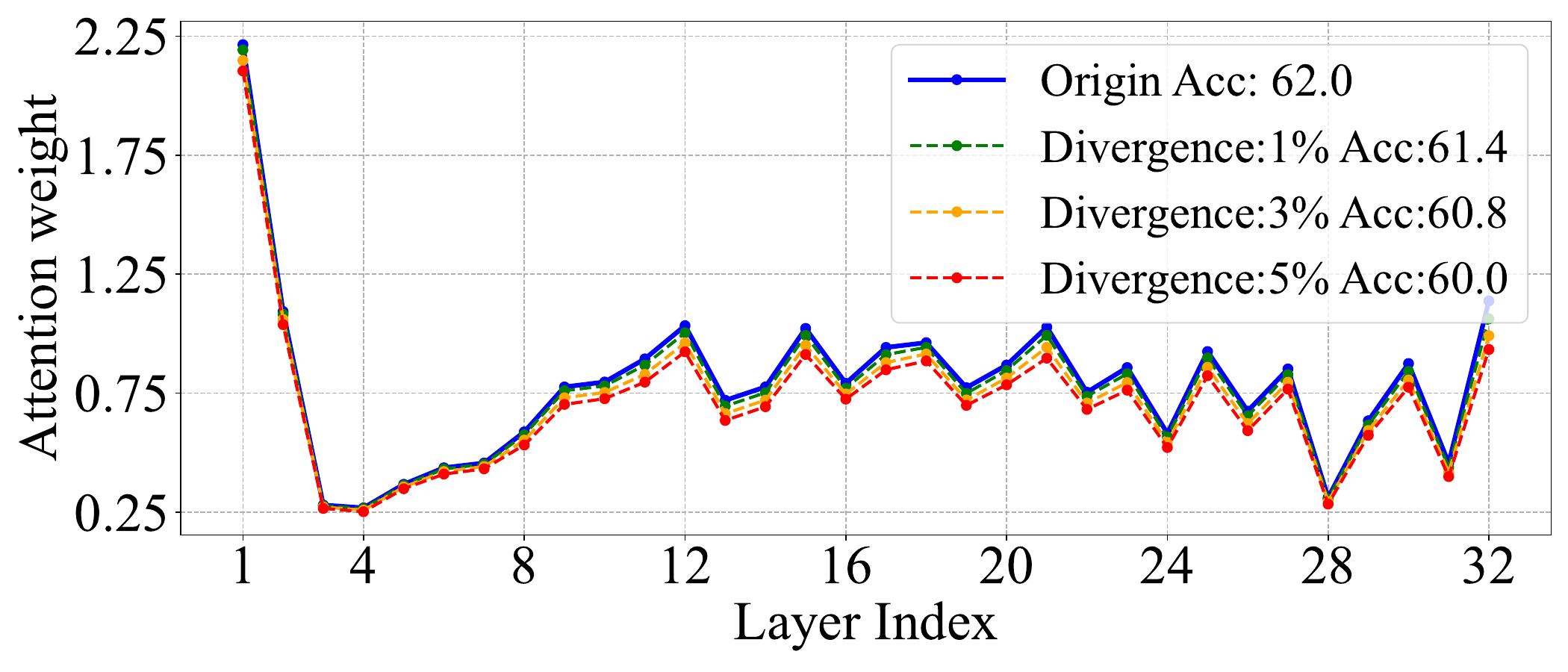}
}
\hspace{-0.015\textwidth} 
\subfigure[Self-Attention Weight Distribution]{
\label{Fig.moti2_self_distribution}
\includegraphics[width=0.491\textwidth]{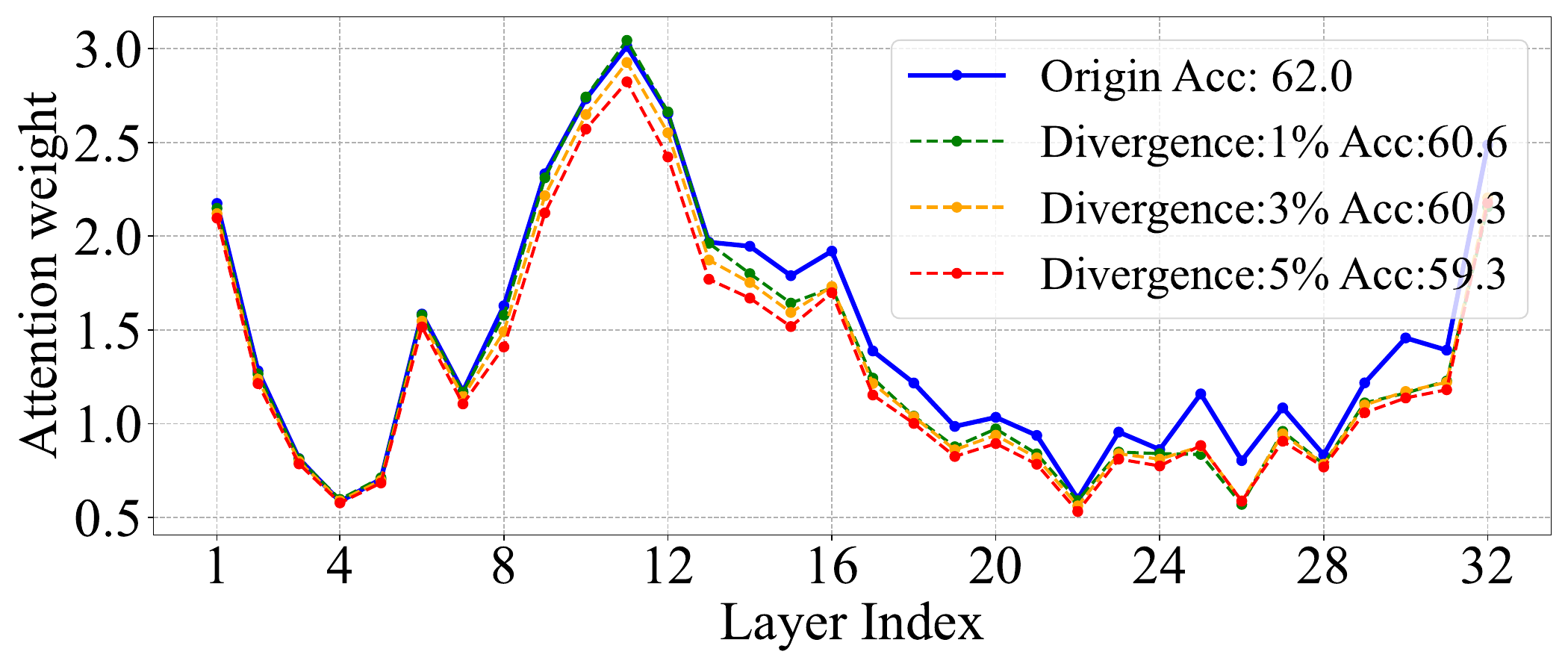}
}
\vspace{-0.4cm}
\caption{ 
The impact of token pruning according to the fitting of cross- and self-attention distributions of visual tokens of LLaVA 1.5. The GQA accuracy \cite{hudson2019gqa} is reported. 
For pruning recipes, the better fitting of attention distribution can retain better performance. However, only considering a single distribution is hard to obtain the optimal pruning recipe. In this paper, our FitPrune will consider the fitting of both cross- and self-attentions.
}
\vspace{-4mm}
\label{fig:moti2}
\end{figure*}

\section{Introduction}

\label{intro}

Recently, the great success of \emph{large language models} (LLMs) ~\cite{zhang2205opt, openai2023gpt, touvron2023llama} also  arouses  an influx of interest in extending them to more modalities, \emph{e.g.}, \emph{vision and language} (VL) \cite{liu2023visual, zhu2023minigpt, wang2023cogvlm,lin2023video, xu2023mplug}. For VL tasks, a simple yet intuitive solution for building \emph{multimodal LLMs} (MLLMs) is to directly project the extracted image features onto the semantic space of LLMs as visual tokens, thus performing multimodal sequence modeling in one Transformer architecture~\cite{liu2023visual, luo2024feast}. 
Despite effectiveness, this paradigm often  suffers from excessive computation. For instance, LLaVA \cite{liu2023visual} uses 576 image patches as the visual tokens, which requires 6.2 times the computation compared to its text-only inference on ScienceQA \cite{lu2022learn}. More recently, practitioners ~\cite{li2023monkey, luo2024feast} resort to increasing image resolution to alleviate the widely criticized visual shortcoming of MLLMs~\cite{tong2024eyes}, which also further exacerbates the already high computation of MLLMs.

In addition, the use of large visual tokens also involves obvious redundancy in MLLMs.  
Compared with previous Transformer-based VL models \cite{zhou2020k,zhou2021trar}, the multi-head attention of MLLMs is unidirectional rather than truly ``global''. Simply put, MLLMs only propagate information from the previous tokens to the subsequent ones, and their visual tokens are usually placed in front of the text  questions. In this case, they mainly serve to give visual semantics to the text tokens, but in fact not many of them are active.
As  shown in Fig.~\ref{fig:moti_1}, the attention from image to text at the 12-\emph{th} layer (32 in total) of LLaVA becomes very focused, indicating that only a small number of visual tokens engage in multimodal reasoning.  Theoretically, pruning the less active tokens would have limited impact on model performance. This assumption is also validated in our experiments, \emph{i.e.} removing a set of visual tokens of MLLMs barely decline the accuracy on most benchmarks. 

However,  when and how to prune the visual tokens of MLLMs  remains an open problem, especially according to a predefined computation budget.
Although token pruning has been extensively studied in \emph{natural language processing} (NLP)~\cite{ye2021tr,liu2023length,anagnostidis2024dynamic} and \emph{computer vision} \cite{kong2022spvit, dong2023heatvit, wei2023joint}, most pruning methods still need to be manually validated for different MLLMs, of which expenditure is also much more expensive. For instance, the researchers often need to define the optimal pruning ratios of  an MLLM and its layers via numerous trials.
The special properties of MLLMs, such as unidirectional self-attention and the involved visual modeling, also make some previous metrics~\cite{fayyaz2022adaptive, wang2023zero} not applicable or suboptimal for VL tasks. For instance, only considering the cross-attention between visual and text tokens will drop the tokens that are also important in visual modeling, and \emph{vice verse}. To this end, we question that

``\emph{Can we find a solution that can directly determine the optimal pruning recipe for MLLMs?}''

To approach this target, we propose an effective and training-free pruning method for MLLMs, termed \emph{Fit and Prune} (FitPrune). 
In particular, we define token pruning as a task of distribution fitting, and adopt statistical principle to obtain the optimal pruning recipe.  Concretely, it aims to minimize the divergence of attention distributions before and after pruning, thereby reducing the negative impact on performance.
In practice, the attention statistics can be well represented by a small batch of data, as shown in Fig. \ref{fig:moti2}. 
Afterwards, FitPrune searches the token slots whose removal would have minimal impact on the default distribution, based on which the pruning ratio of each layer can be directly set. 
To better meet the property of MLLMs, FitPrune also considers the distributions of both cross- and intra-attentions of visual tokens, \emph{i.e.}, the two distributions in Fig.\ref{fig:moti2}.  With these innovative designs, FitPrune can directly  produce a pruning recipe  according to a predefined computation budget. 

To validate FitPrune, we apply it to a set of advanced MLLMs, including LLaVA-1.5~\cite{liu2023improved}, LLaVA-HR~\cite{luo2024feast} and LLaVA-NEXT~\cite{liu2024improved}, on  a bunch of highly competitive VL benchmarks~\cite{lu2022learn,singh2019towards,hudson2019gqa,conf/cvpr/GoyalKSBP17,gurari2018vizwiz, fu2023mme, liu2023mmbench,yu2023mmvet,li2023evaluatingpope}. Experiment results show that FitPrune can significantly reduce the computational complexity of MLLMs while retaining high performance on all benchmarks. For instance, FitPrune can reduce 54.9\% TFLOPs of LLaVA-NEXT with only 0.5\% performance degradation.
More importantly, its pruning recipe can be obtained in about 5 minutes for all VL tasks. 


Conclusively, the contribution of this paper is three-fold:
\begin{itemize}
\item  We reveal the heavy redundancy of visual tokens of MLLMs via investigating their attention patterns, indicating that a great number of visual tokens can be actually discarded during inference.

\item  To achieve fast and effective token pruning of MLLMs, we propose  a novel and training-free  method called \emph{fit-and-prune (FitPrune)}, which consider token pruning as a task of cross-distribution fitting. 

\item On ten VL benchmarks, FitPrune not only significantly reduces the computation overhead, e.g., -54.1\% TFLOPs of LLaVA-NEXT with marginal performance decline, but also achieves a better trade-off than existing visual token pruning methods.

\end{itemize}

\begin{figure*}[!th]
\centering
\includegraphics[width=1.0\linewidth]{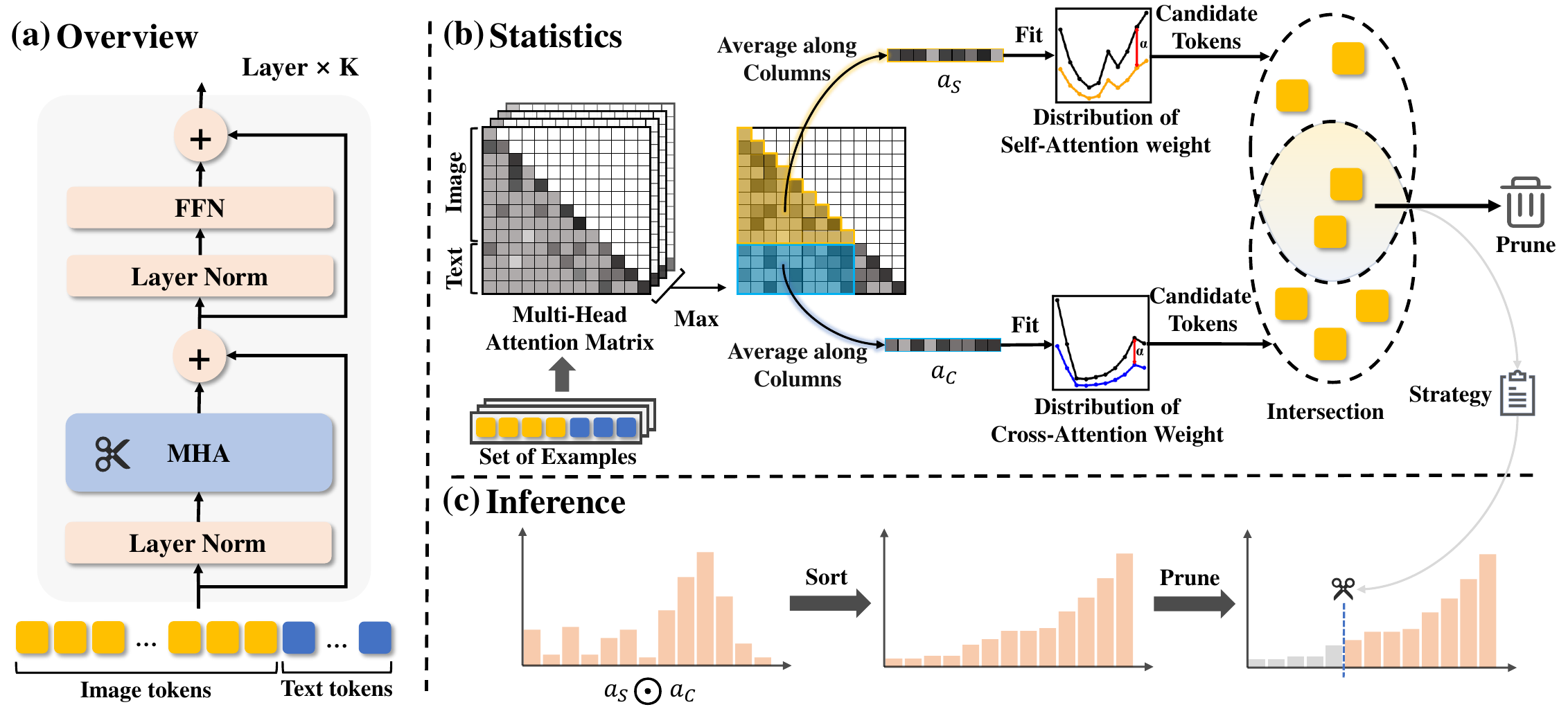}
\vspace{-0.3cm}
\caption{ Illustration of our FitPrune. 
(a) FitPrune is used to reduce the length of visual tokens in the MHA of each layer. (b) The generated pruning recipe is obtained via binary search based on the attention statistics of a set of examples. Its principle is to find out the optimal pruning recipe that  reduce the gap of distributions before and after pruning. (c) During inference, the MLLM can drop tokens according to the pruning recipe of FitPrune. 
}

\vspace{-4mm}
\label{Fig_1_main}
\end{figure*}

\section{Related Work}

\subsection{Vision-Language Models}
Recently, the prevalence of \emph{large language models} (LLMs) leads to a trend of extending these LLMs to VL tasks ~\cite{touvron2023llama, le2023bloom}, termed \emph{multimodal LLMs}  (MLLMs) or \emph{vision-LLMs} (VLLMs).
Compared with previous single-task or BERT-style VLMs~\cite{radford2021learning, bao2022vlmo}, MLLMs ~\cite{liu2023visual,liu2024improved} differ in the direct use of language models for both language modeling and multimodal interactions, rather than building another deep multimodal branch.
A main design paradigm of these MLLMs is to use a cross-modal projector to connect the visual encoder~\cite{bai2023qwen,liu2023visual} with LLMs \cite{touvron2023llama, le2023bloom}, which can be a deep Query-based branch \cite{bai2023qwen, li2023blip} or a simple  projection network ~\cite{luo2024feast,liu2024improved}. Despite effectiveness, existing MLLMs still suffer from visual shortcoming~\cite{tong2024eyes} or visual hallucination~\cite{huang2024visual} mainly due to the inferior descriptive power of input visual tokens \cite{liu2023hidden}. To this end, recent efforts~\cite{li2023monkey, luo2024feast} have been devoted to increasing the image resolution for MLLMs, which exhibit obvious improvements in fine-grained VL tasks, such as TextVQA \cite{singh2019towards}. However, the increase of image resolution also greatly exacerbate the already high computation of MLLMs.  
Meanwhile, the large visual tokens also involve obvious redundancy as discussed in this paper. 
In this case, how to effectively and adaptively prune less important tokens of MLLM is a critical demand for its applications.


\subsection{Token Pruning}
Token pruning, as a subclass of \emph{dynamic neural networks} \cite{han2021dynamic}, has become a hot research topic with the prevalence of Transformer-based networks~\cite{vaswani2017attention, devlin2018bert, dosovitskiy2020image}, aiming to speed up inference by dynamically reducing less important tokens based on input-dependent importance.
This task has been actively investigated in either \emph{natural language processing} (NLP)\cite{ye2021tr, kim2022learned} or \emph{computer vision} (CV) \cite{rao2021dynamicvit, meng2022adavit}. Existing methods~\cite{rao2021dynamicvit, meng2022adavit, kong2022spvit,wei2023joint,dong2023heatvit} mainly focus on the design of pruning metrics, \emph{e.g.}, low-information, and often require post-pruning tuning. It is prohibitively expensive for LLMs and MLLMs. For example, DiffRate~\cite{chen2023diffrate} relies on prediction loss to guide pruning and requires a large dataset for optimization, while FitPrune uses only a small batch of data, leveraging attention distribution fitting before and after pruning. 
Some recent approaches explore training-free token pruning~\cite{fayyaz2022adaptive, bolya2022token, wang2023zero, chen2024image}. For instance, ATS~\cite{fayyaz2022adaptive}  adopts token sampling based on importance scores, ToMe~\cite{bolya2022token} gradually merges tokens by similarity, and Zero-TPrune~\cite{wang2023zero} employs a  weighted  \emph{PageRank} algorithm to identify less critical tokens. In the visual-language field, token pruning is nascent. MADTP~\cite{cao2024madtp} employs a novel  multimodal alignment guidance module coupled with \emph{dynamic token pruning} to efficiently compress the model. FastV~\cite{chen2024image} introduces a pruning strategy that determines token importance based on average attention received by each token. 
Despite effectiveness, existing methods still need  to manually determine the pruning ratio of Transformer and its layers, which makes trials expensive for MLLMs. In this paper, we focus on a training-free and cost-effective method to quickly determine the pruning recipe based on the predefined target.



\section{Method}
In this paper, we propose a novel and training-free method towards fast and effective visual token pruning of \emph{multimodal large language models} (MLLMs), termed \emph{fit-and-prune} (FitPrune). The main principle of FitPrune is to remove the redundant visual tokens based on the fit  of attention distributions before and after pruning. 
In practice, FitPrune performs statistical analysis to derive pruning strategies, as illustrated in Fig. \ref{Fig_1_main}. 

Concretely, given a model $G(\cdot)$ and  its averaged  attention distribution \( D \) on a set of examples, the objective of FitPrune is to find a strategy \( P \) that minimizes the divergence between $D$ and the pruned attention distribution $D'$ given a computation budget \( \delta \):

\begin{equation}
\arg\min_{P} \  d(D, D') 
\quad \text{s.t.} \quad \Phi\left( G,  P \right) \leq {\delta},
\label{func1}
\end{equation}
\noindent  where $\Phi(G, P)$ denotes the computation overhead of $G (\cdot)$ under the pruning strategy $P$, and $\delta$ is the given computation budget. In this paper, we primarily focus on FLOPs.


To meet the property of MLLMs, FitPrune considers the self- and cross-attentions at the same time, denoted as $D_S=[\bar{a}_s^i]_{i=1}^K$ and $D_C=[\bar{a}_c^i]_{i=1}^K$, respectively. 
Here, $K$ is the number of layers, and $\bar{a}_s^i$ denotes the averaged self-attention value of visual tokens at the $i$-\emph{th} layer:

\begin{equation}
\bar{a}_s^i=\frac{1}{N}\sum_{m=1}^{N}\sum_{n=1}^{N} A^{i}_{m,n} ,
\label{importanceofs}
\end{equation}



\noindent where \( A^i \in \mathbb{R}^{(N+M) \times (N+M)} \) represents the attention matrix of  the $i$-\emph{th} layer, and \( N \) and \( M \) denote the numbers of visual and text tokens, respectively. In practice, $A$ is the averaged statistics of a batch of examples, akin with $D$. 

Similarly, $\bar{a}_c^i$ is the averaged cross-attention value of visual tokens, obtained by

\begin{equation}
\bar{a}_c^i= \frac{1}{M}\sum_{m=N+1}^{N+M} \sum_{n=1}^{N} A^{i}_{m,n}.    
\label{importanceofc}
\end{equation}



\noindent Afterwards, we denote these two distributions after pruning as 
\(D'_S = [\bar{a'}_s^{i}]_{i=1}^K\) and \(D'_C = [\bar{a'}_c^{i}]_{i=1}^K\), respectively.
Here, $\bar{a'}_s^i$ and $\bar{a'}_c^i$ are obtained by:



\begin{equation}
\bar{a'}_s^i=\frac{1}{N - t_i}\sum_{m=1}^{N - t_i}\sum_{n=1}^{N  - t_i}{A'}^{i}_{m,n} ,
\end{equation}

\begin{equation}
\bar{a'}_c^i= \frac{1}{M}\sum_{m=N-t_i+1}^{N+M-t_i}\sum_{n=1}^{N - t_i}{A'}^{i}_{m,n}. 
\end{equation}



\noindent where \({A'}^i \in \mathbb{R}^{(N + M - t_i) \times (N + M - t_i)}\) is the attention matrix after pruning.
Since the number of removed tokens increases layer by layer, $t_i$ is a cumulative value, defined by $t_i=\sum_{i-1}t_i+t_i^*$, and $t_i^*$ is the number of new tokens removed at the $i$-\emph{th} layer.
In this case, the final pruning strategy $P$ can be denoted by

\begin{equation}
P = [t_1^*, t_2^*, \cdots, t_K^*].
\end{equation}


Lastly, the optimization of FitPrune in Eq.\ref{func1} can be reformulated as

\begin{equation}
\begin{aligned}
& \arg\min_{P} \  \left[ d(D_S, {D'}_S) + d(D_C, {D'}_C) \right], \\
& \text{s.t.} \quad \Phi\left( G, P ) \right) \leq {\delta}.
\end{aligned}
\label{func:objective_batch}
\end{equation}

\begin{table*}[!th]
\resizebox{\textwidth}{!}{%
\vspace{1mm}
\setlength{\tabcolsep}{0.4mm}
\renewcommand{\arraystretch}{1.1}
\begin{tabular}{l c |c c|c c|c c|c c|c c|c c}
\toprule 
\multirow{2}{*}{\makecell[c]{\textbf{Method}} } 
& \multirow{2}{*}{\makecell[c]{\textbf{Pruning}\\\textbf{Ratio}}} 
& \multicolumn{2}{c|}{ \textbf{ScienceQA-IMG} } 
& \multicolumn{2}{c|}{ \textbf{GQA} } 
& \multicolumn{2}{c|}{ \textbf{TextVQA} } 
& \multicolumn{2}{c|}{ \textbf{Vizwiz} } 
& \multicolumn{2}{c|}{ \textbf{VQAv2} } 
& \multicolumn{2}{c}{ \textbf{Average} }\\

\cmidrule{3-14} 
& & \textbf{Accuracy} & \textbf{TFLOPs*} & \textbf{Accuracy} & \textbf{TFLOPs*} & \textbf{Accuracy} & \textbf{TFLOPs*} & \textbf{Accuracy} & \textbf{TFLOPs*} & \textbf{Accuracy} & \textbf{TFLOPs*} & \textbf{Accuracy} & \textbf{TFLOPs*} \\

\midrule 

\rowcolor{gray!20} LLaVA-1.5 7B & 0\% & 68.0 & 10.0 & 62.0 & 9.1 & 58.2 & 9.9 & 50.0 & 9.3 & 78.5 & 9.1 & 63.3 & 9.5 \\

+FitPrune & 40\% & 67.7 & 7.0 ($-30.5\%$) & 61.9 & 6.0 ($-33.7\%$) & 58.1 & 6.8 ($-31.3\%$) & 50.2 & 6.2 ($-33.0\%$) & 78.5 & 6.0 ($-33.9\%$) & 63.3 & 6.4($-32.6\%$) \\

+FitPrune & 60\% & 67.8 & 5.3 ($-46.8\%$) & 61.5 & 4.4 ($-51.6\%$) & 58.2 & 5.1 ($-47.9\%$) & 50.4 & 4.6 ($-50.5\%$) & 78.3 & 4.4 ($-51.9\%$) & 63.2 & 4.8($-49.5\%$) \\

\midrule 

\rowcolor{gray!20} LLaVA-HR 7B & 0\% & 68.0 & 18.2 & 64.2 & 17.3 & 67.1 & 18.0 & 48.7 & 17.5 & 81.9 & 17.2 & 66.0 & 17.6 \\

+FitPrune & 40\% & 68.0 & 12.7 ($-30.6\%$) & 64.2 & 11.7 ($-32.1\%$) & 67.2 & 12.5 ($-30.7\%$) & 48.5 & 11.9 ($-31.7\%$) & 81.9 & 11.7 ($-32.1\%$) & 66.0 & 12.1($-31.3\%$) \\

+FitPrune & 60\% & 67.7 & 9.7 ($-46.7\%$) & 64.1 & 8.8 ($-49.1\%$) & 66.7 & 9.5 ($-47.1\%$) & 48.5 & 9.0 ($-48.5\%$) & 81.8 & 8.7 ($-49.2\%$) & 65.8 & 9.1($-48.3\%$) \\

\midrule 

\rowcolor{gray!20} LLaVA-NEXT 7B & 0\% & 70.2 & 26.1 & 64.2 & 35.2 & 64.9 & 37.3 & 57.7 & 37.0 & 81.8 & 35.1 & 67.1 & 34.1 \\

+FitPrune & 40\% & 70.1 & 17.2 ($-34.3\%$) & 64.2 & 22.3 ($-36.6\%$) & 64.9 & 23.8 ($-36.0\%$) & 57.4 & 23.5 ($-36.5\%$) & 81.7 & 22.2 ($-36.6\%$) & 66.9 & 21.8($-36.1\%$) \\

+FitPrune & 60\% & 70.1 & 12.6 ($-51.7\%$) & 64.0 & 15.9 ($-55.0\%$) & 64.2 & 17.1 ($-54.0\%$) & 57.3 & 16.7 ($-54.9\%$) & 81.5 & 15.8 ($-55.0\%$) & 66.8 & 15.6($-54.3\%$) \\

\bottomrule[1.00pt]
\end{tabular}
}
\vspace{-3mm}
\caption{
Results of MLLMs with FitPrune on 5 MLLM benchmarks. 
Pruning ratio denotes the target reduction of FLOPs by visual tokens. TFLOPs* reflect the computation and actual FLOPs reduction of the entire MLLM. Due to the different lengths of text decoding, the overall reduction varies for different benchmarks.
}
\label{Tab:main}
\vspace{-2mm}
\end{table*}

\begin{table*}[!t]
%
\resizebox{\textwidth}{!}{

\vspace{1mm}
\setlength{\tabcolsep}{0.4mm}
\renewcommand{\arraystretch}{1.1}
\begin{tabular}{l c |c c|c c|c c|c c|c c|c c}

\toprule 
\multirow{2}{*}{\makecell[c]{\textbf{Method}} } 
& \multirow{2}{*}{\makecell[c]{\textbf{Pruning}\\\textbf{Ratio}}} 
& \multicolumn{2}{c|}{ \textbf{POPE} } 
& \multicolumn{2}{c|}{ \textbf{MM-Vet} } 
& \multicolumn{2}{c|}{ \textbf{MMB} } 
& \multicolumn{2}{c|}{ \textbf{ MMB$^{\textnormal{CN}}$  } } 
& \multicolumn{2}{c|}{ \textbf{MME} }
& \multicolumn{2}{c}{ \textbf{Average} } \\

\cmidrule{3-14} 
& & \textbf{Accuracy} & \textbf{TFLOPs*} & \textbf{Accuracy} & \textbf{TFLOPs*} & \textbf{Accuracy} & \textbf{TFLOPs*} &\textbf{ Accuracy }&\textbf{ TFLOPs* }& \textbf{ Accuracy }&\textbf{ TFLOPs* } & \textbf{Accuracy} & \textbf{TFLOPs*} \\

\midrule 

\rowcolor{gray!20} LLaVA-1.5 7B & 0\% & 85.9 & 9.1 & 31.6 & 10.0 & 64.3 & 9.8 & 58.3 & 10.4 & 1510.7 & 9.1 & 63.1 & 9.7 \\

+FitPrune & 40\% & 86.9 & 6.0($-33.9\%$) & 31.5 & 6.9($-30.7\%$) & 64.9 & 6.8($-31.3\%$) & 58.2 & 7.3($-29.8\%$) & 1502.5 & 6.1($-33.5\%$) & 63.4 & 6.6($-32.0\%$) \\

+FitPrune & 60\% & 86.5 & 4.4($-51.8\%$) & 32.8 & 5.3($-47.1\%$) & 64.6 & 5.1($-47.9\%$) & 58.4 & 5.6($-45.7\%$) & 1507.9 & 4.4($-51.4\%$) & 63.5 & 5.0($-48.5\%$) \\

\midrule 

\rowcolor{gray!20} LLaVA-HR 7B & 0\% & 87.6 & 17.2 & 31.2 & 17.9 & 65.0 & 18.1 & 60.6 & 18.6 & 1554.9 & 17.3 & 64.4 & 17.8\\

+FitPrune & 40\% & 88.4 & 11.7($-32.2\%$) & 31.5 & 12.3($-31.1\%$) & 65.0 & 12.5($-30.7\%$) & 60.1 & 13.0($-30.1\%$) & 1543.9 & 11.7($-32.1\%$) & 64.4 & 12.2($-31.5\%$) \\

+FitPrune & 60\% & 88.7 & 8.7($-49.2\%$) & 31.0 & 9.4($-47.5\%$) & 64.9 & 9.6($-47.0\%$) & 60.2 & 10.0($-46.0\%$) & 1561.6 & 8.8($-49.0\%$) & 64.6 & 9.3($-47.8\%$) \\

\midrule 

\rowcolor{gray!20} LLaVA-NEXT 7B & 0\% & 86.5 & 35.0 & 43.9 & 37.3 & 67.4 & 31.1 & 60.6 & 31.6 & 1519.0 & 33.2 & 66.9 & 33.6 \\

+FitPrune & 40\% & 87.6 & 22.2($-36.5\%$) & 44.2 & 24.5($-34.3\%$) & 68.0 & 20.0($-35.5\%$) & 60.3 & 20.6($-35.0\%$) & 1506.0 & 21.1($-36.5\%$) & 67.1 & 21.7($-35.4\%$) \\

+FitPrune & 60\% & 87.6 & 15.8($-54.8\%$) & 41.7 & 17.8($-52.2\%$) & 67.5 & 14.5($-53.4\%$) & 60.0 & 15.0($-52.6\%$) & 1486.0 & 15.0($-54.8\%$) & 66.2 & 15.0($-55.4\%$) \\

\bottomrule[1.00pt]
\end{tabular}
}

\vspace{-3mm}
\caption{
Performance with different pruning ratios on three MLLMs for 5 multimodal benchmarks for MLLMs. 
Pruning ratio denotes the target reduction of FLOPs by visual tokens. TFLOPs* reflect the computation and actual reduction of FLOPs of the entire MLLM.}

\label{Tab:main_additional}
\vspace{-8mm}
\end{table*}

\begin{algorithm}[!t]
\caption{FitPrune}
\label{algorithm_RF}
\begin{algorithmic}[1]
\REQUIRE model $G$,  target FLOPs budget \( \delta \) and binary search threshold \( \epsilon \)
\ENSURE Pruning strategy \(P\)

\STATE  Initialize \ $
            D_S, \
            D_C
        $
\STATE Initialize \( \alpha_{\text{L}} = 0 \), \( \alpha_{\text{R}} = 1 \), \( P \)

\WHILE{\( \alpha_{\text{R}} - \alpha_{\text{L}} > \epsilon  \)}
    \STATE \( \alpha = (\alpha_{\text{L}} + \alpha_{\text{R}}) / 2 \)
        \FOR{each layer \( i \) in model \( G \)}
            \STATE Initialize the candidate sets \( \mathcal{T}_{S} = \emptyset \), \( \mathcal{T}_{C} = \emptyset \)
            \STATE Compute \(a_s^{i}\) and \(a_c^{i}\) according to Eq.\ref{importances}
            \STATE \( I_S = \text{\textit{argsort}}(a_s^{i}) \), \( I_C = \text{\textit{argsort}}(a_c^{i}) \)
            \STATE \( \mathcal{S} = \{(I_{S}, \bar{a}_{s}^{i}, {a}_{s}^{i}, \mathcal{T}_{S}), (I_{C}, \bar{a}_{c}^{i},{a}_{c}^{i}, \mathcal{T}_{C})\} \)

            \FOR{each \( (I, \bar{a}^i, {a}^i, \mathcal{T}) \) in \( \mathcal{S} \)}
                \FOR{each \(k\) in \(I\)}
                    \STATE \( \mathcal{T} = \mathcal{T} \cup \{k\} \)
                    \IF{ \( \left| \frac{ \bar{a}^i - \sum_{k \in \mathcal{T}} a^{i,k} }{\bar{a}^i} \right| > \alpha \)}
                        \STATE \( \mathcal{T} = \mathcal{T} \setminus \{k\} \)
                        \STATE \textbf{break}
                    \ENDIF
                \ENDFOR
            \ENDFOR
            \STATE \( t_{i}^* = |\mathcal{T_{S}} \cap \mathcal{T_{C}}| \)
        \ENDFOR
    \IF{ \( \Phi\left( G(P, x) \right) \leq \delta \)}
        \STATE \( \alpha_{\text{R}} = \alpha \)
    \ELSE
        \STATE \( \alpha_{\text{L}} = \alpha \)
    \ENDIF
\ENDWHILE
\RETURN strategy $P=[t_1^*, t_2^*, \cdots, t_K^*]$
\end{algorithmic}
\end{algorithm}

To accomplish the objective of Eq.\ref{func:objective_batch}, we adopt the principle of \emph{binary search} to obtain the optimal pruning recipe. The search algorithm is depicted in Algorithm \ref{algorithm_RF}.
Concretely, we first obtain the default the self-attention distribution $D_S = [ a_{s}^{i} ]_{i=1}^K$ and cross-attention distribution $D_C = [ a_{c}^{i} ]_{i=1}^K$ from a set of examples. 
Then, we employ binary search to determine the smallest divergence upper bound \(\alpha\). The binary search is initialized with \(\alpha_{\text{L}} = 0\) as the left boundary (no divergence) and \(\alpha_{\text{R}} = 1\) as the right boundary (maximum divergence). In each iteration of the binary search, the midpoint \(\alpha = (\alpha_{\text{L}} + \alpha_{\text{R}}) / 2\) is evaluated. For each candidate value of \(\alpha\), we greedily remove the token that has the smallest impact on the attention distributions.


Concretely, for the \(i\)-\emph{th} layer, we first evaluate the received attention weights for each visual token. The \(j\)-th token's attention weights in two distributions are denoted as \(a_s^{i} \in \mathbb{R}^{N-t_{i-1}}\) and \(a_c^{i} \in \mathbb{R}^{N-t_{i-1}}\), calculated by:

\noindent\vspace{-6mm}

\begin{equation}
a_s^{i,j} = \sum_{m=1}^{N - t_{i-1}} A^{i}_{m,j}, \quad
{a}_c^{i,j} = \sum_{m=N-t_{i-1}+1}^{N+M-t_{i-1}} A^{i}_{m,j}.
\label{importances}
\end{equation}


Next, we select the sets of tokens \(\mathcal{T}_S\) and \(\mathcal{T}_C\) with the smallest attention weights in the self-attention and cross-attention distributions, so that their removal results in a distribution divergence close to \(\alpha\):

\noindent\vspace{-4mm}

\begin{equation}
\mathcal{T}_S = \operatorname*{argmax}_{\mathcal{T} \subseteq [N - t_{i-1}]} \left| \mathcal{T} \right| \; \text{s.t.} \; \left| \frac{ \bar{a}_{s}^i - \sum_{j \in \mathcal{T}} a_s^{i,j} }{\bar{a}_{s}^i}\right| \leq \alpha,   
\end{equation}

\begin{equation}
\mathcal{T}_C = \operatorname*{argmax}_{\mathcal{T} \subseteq [N - t_{i-1}]} \left| \mathcal{T} \right| \; \text{s.t.} \; \left|\frac{ \bar{a}_{c}^i - \sum_{j \in \mathcal{T}} a_c^{i,j} }{\bar{a}_{c}^i}\right| \leq \alpha,
\end{equation}

\noindent where \(\mathcal{T}\) represents a subset of the remaining tokens. The goal is to maximize the number of tokens \(|\mathcal{T}|\) that can be removed while ensuring that the divergence between the original and pruned attention distributions does not exceed the divergence threshold \(\alpha\).
Then, the number of pruned tokens in $i$-\emph{th} layer can be calculated by:

\begin{equation}
t_i^* = |\mathcal{T}_i|,\quad \text{where} \;
\mathcal{T}_i = \mathcal{T}_S \cap \mathcal{T}_C.
\end{equation}



This iterative process continues by checking whether \(\Phi(G, P)\) meets the target computation budget \( \delta \) after each iteration, and accordingly adjusting the binary search boundaries \(\alpha_{\text{L}}\) and \(\alpha_{\text{R}}\). The search terminates when the difference between \(\alpha_{\text{R}}\) and \(\alpha_{\text{L}}\) is no more than the threshold \(\epsilon\).

\begin{figure}[!t]
    \centering
    \includegraphics[width=1.0\linewidth]{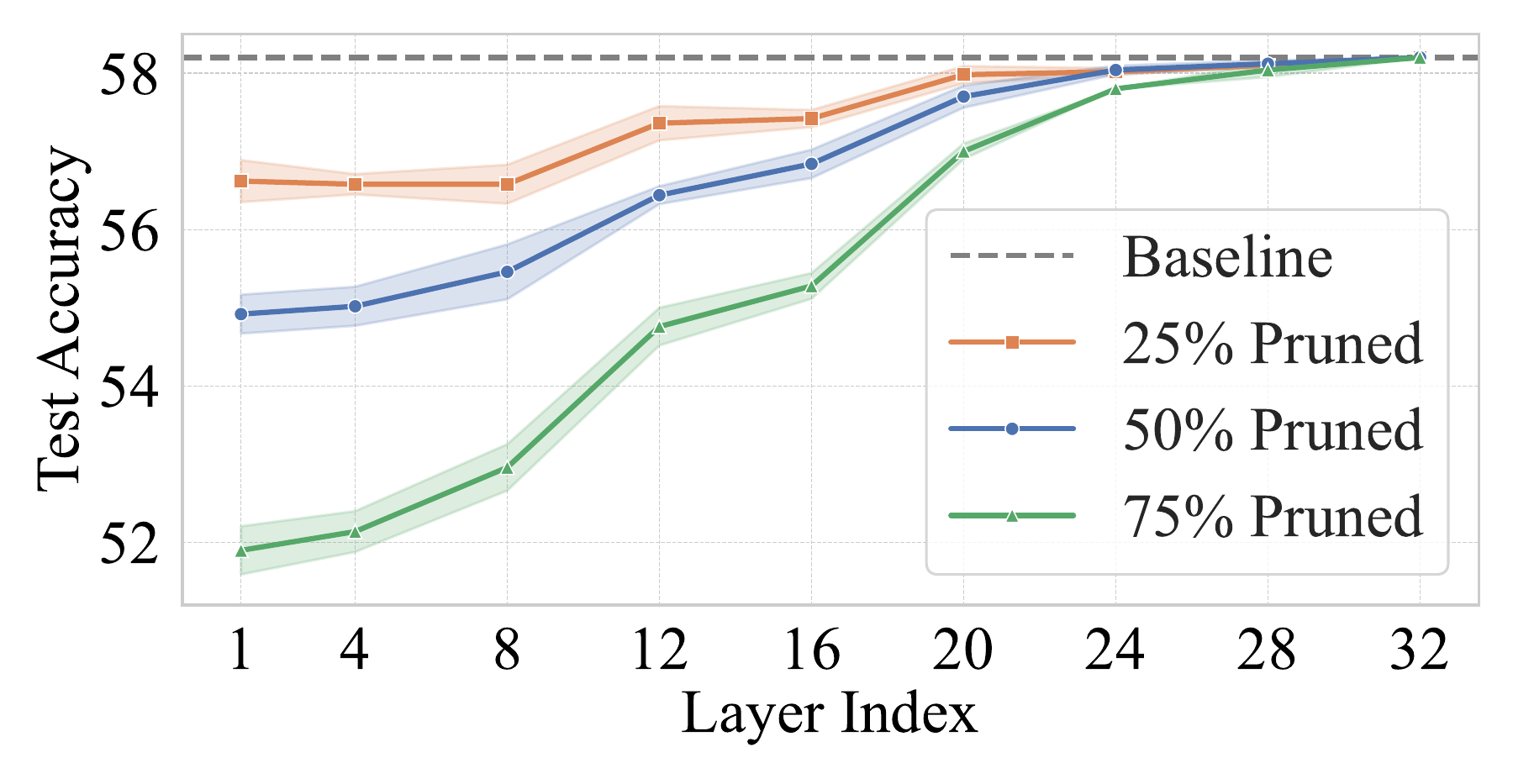}
    \vspace{-8mm}
    \caption{
    Performance of LLaVA-1.5 using different ratios of random pruning on TextVQA.
    }
    \vspace{-6mm}
    \label{fig:token_redundancy_analysis}
\end{figure}

During inference, tokens are ranked by the combined importance metric 
$a_u^{i,j} = a_s^{i,j} \cdot a_c^{i,j} $,  where \(a_s^{i,j}\) and \(a_c^{i,j}\) are the attention weights received by the \(j\)-th image token in the $i$-\emph{th} layer from other image tokens and the text tokens , respectively.
Tokens with the lowest \(a_u^{i,j}\) are pruned first. The tokens to be pruned in \(i\)-\emph{th} layer can be defined as:


\begin{equation}
\mathcal{T}_i = \operatorname*{argmin}_{\mathcal{T} \subseteq [N - t_{i-1}], |\mathcal{T}| = t_i^*} \; \sum_{j \in \mathcal{T}} a_u^{i,j}
\end{equation}

\noindent where \(\mathcal{T}\) represents a subset of the remaining tokens, and \( t_i^* \) represents the number of tokens to be pruned from layer \( i \) as specified in the pruning strategy \( P \). This ensures that the least important tokens are removed, preserving model performance while enhancing computational efficiency. FitPrune does not need any additional modules and also requires no gradient computations, thus ensuring its efficient deployment with minimal computational overhead.

\begin{figure*}[!ht]
\centering  
\subfigure[VQAv2]{
\label{Fig.compare_gqa2}
\includegraphics[width=0.318\textwidth]{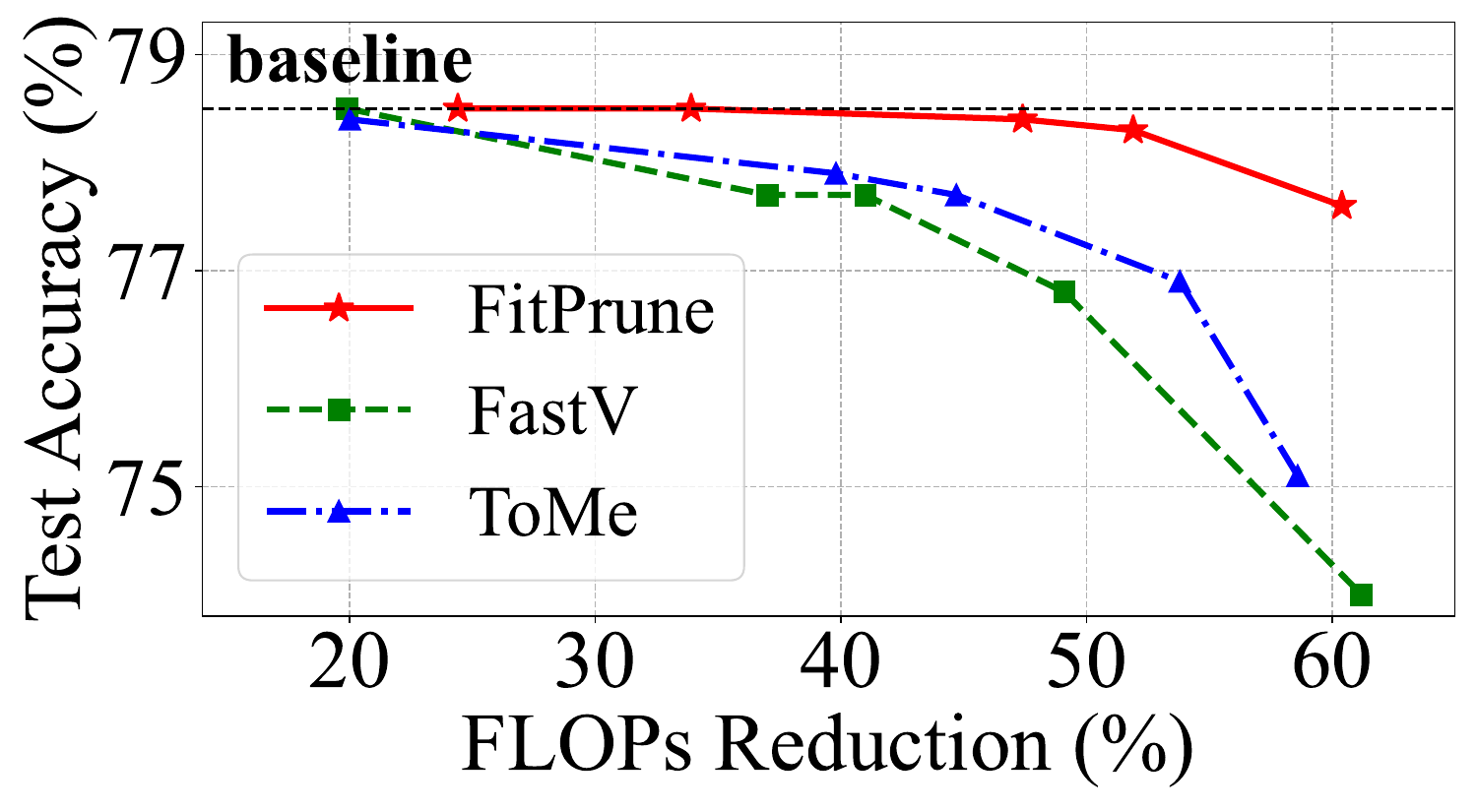}
}
\hspace{-0.015\textwidth} 
\subfigure[GQA]{
\label{Fig.compare_gqa1}
\includegraphics[width=0.325\textwidth]{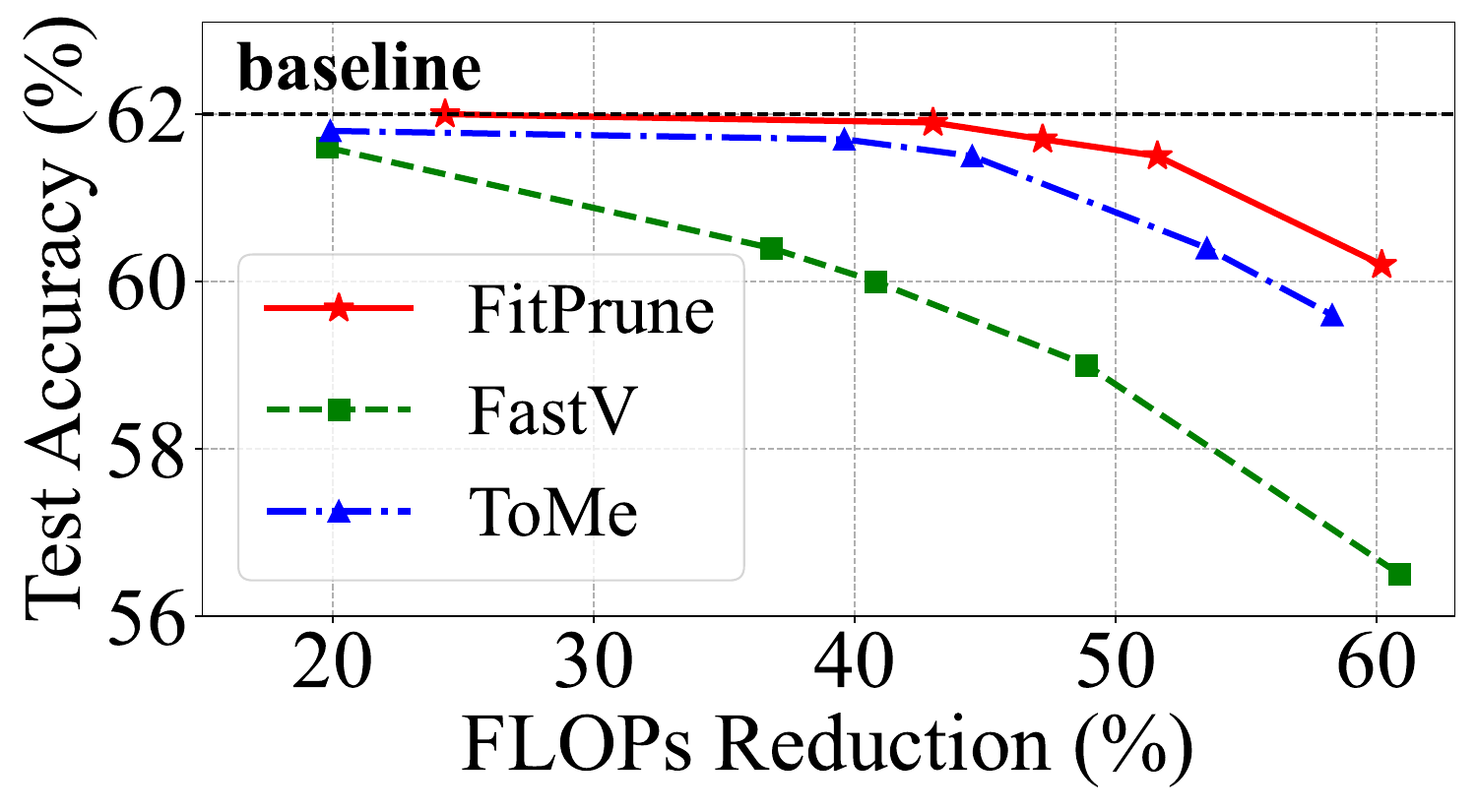}
}
\hspace{-0.015\textwidth} 
\subfigure[TextVQA]{
\label{Fig.compare_textvqa}
\includegraphics[width=0.325\textwidth]{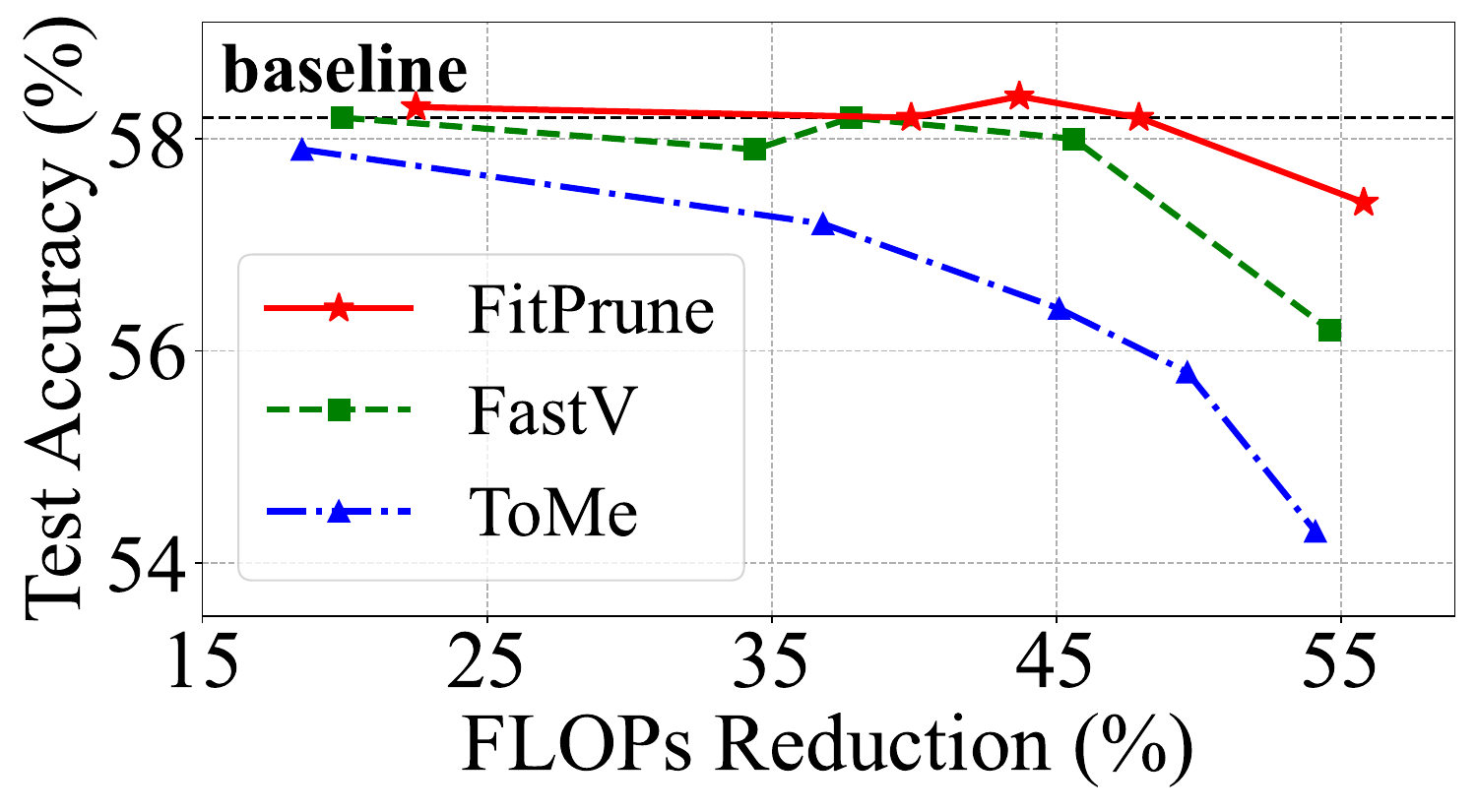}
}
\vspace{-5mm}
\caption{
Performance comparison of  FitPrune and other pruning methods on the LLaVA-1.5 7B \emph{w.r.t} different pruning ratios.
}
\vspace{-4mm}
\label{fig:comparison}
\end{figure*}

\begin{figure*}[!t]
\centering  
\subfigure[Distribution of Cross-attention Weight]{
\label{d_cross}
\includegraphics[width=0.325\textwidth]{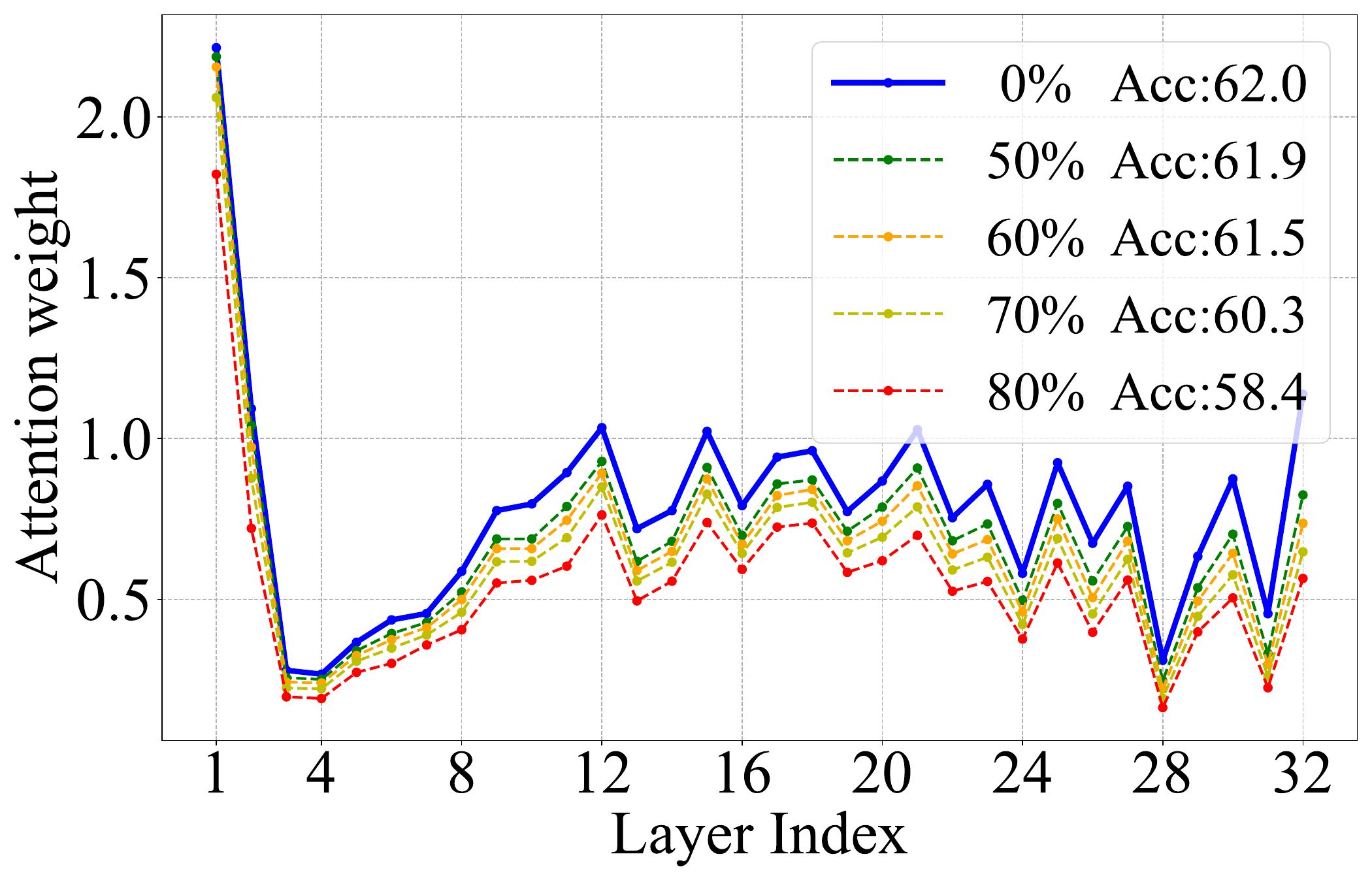}
}
\hspace{-0.015\textwidth} 
\subfigure[Distribution of Self-attention Weight]{
\label{d_self}
\includegraphics[width=0.325\textwidth]{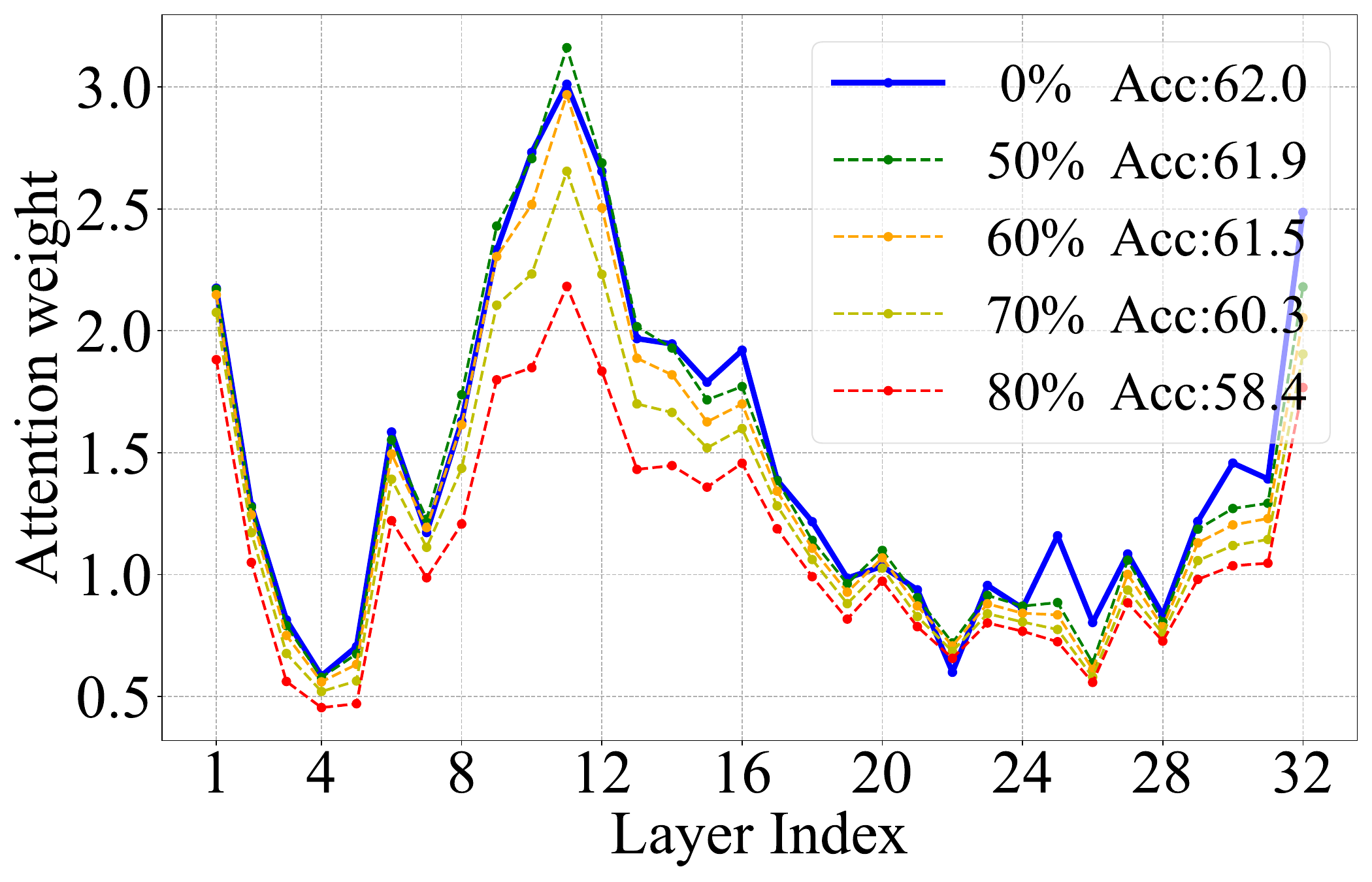}
}
\hspace{-0.015\textwidth} 
\subfigure[Strategy Visualization]{
\label{aaa}
\includegraphics[width=0.325\textwidth]{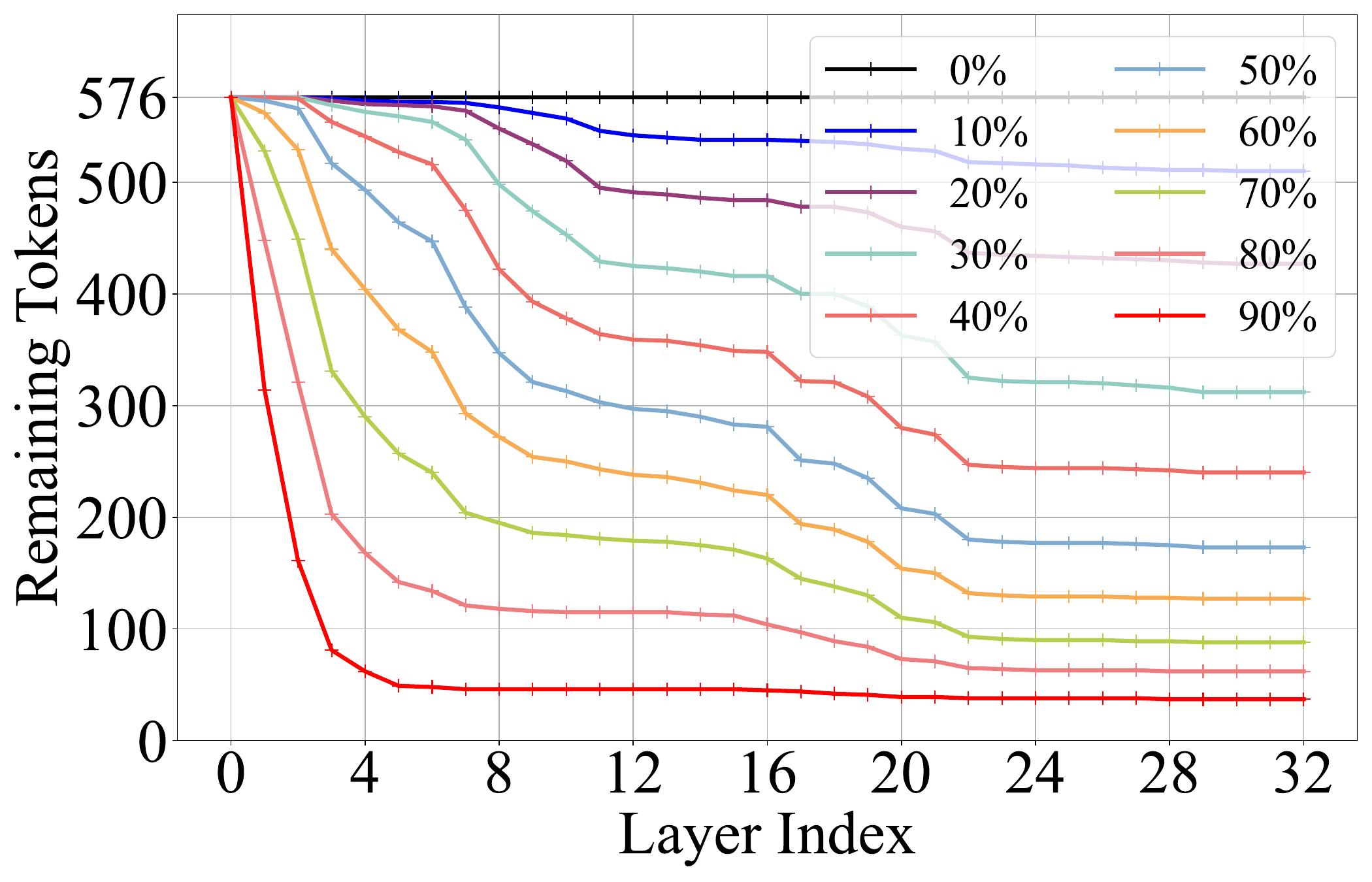}
}
\vspace{-5mm}
\caption{
The visualization of impact of our FitPrune.
(a) Distribution of cross-attention weights.
(b) Distribution of self-attention weights.
These distributions are based on LLaVA-1.5 7B for the GQA benchmark.
(c) The number of tokens across the 32 layers under the different settings.
}
\vspace{-5 mm}
\label{fig:abla}
\end{figure*}

\vspace{-1mm}
\section{Experiments}
\subsection{Datasets and Metrics}
To validate FitPrune, we conduct  extensive experiments on 10 benchmarks. The datasets include the common vision-language benchmarks such as  VQAv2~\cite{conf/cvpr/GoyalKSBP17} for visual question answering, GQA~\cite{hudson2019gqa} for compositional reasoning, VizWiz~\cite{gurari2018vizwiz} for zero-shot generalization, ScienceQA-IMG~\cite{lu2022learn} for scientific question answering, and TextVQA~\cite{singh2019towards} for text understanding in images.
The multimodal benchmarks for MLLMs are also used, including POPE~\cite{li2023evaluatingpope} for object hallucination, MM-Vet~\cite{yu2023mmvet} for integrated vision-language capabilities, MMBench~\cite{liu2023mmbench} for diverse perception and reasoning tasks, MMBench$^{\text{CN}}$ for Chinese language evaluation, and MME~\cite{fu2023mme} for systematic multimodal assessment. All experiments on these datasets follow the default setting and metric.

\vspace{-2mm}

\subsection{Implementation Details}

To conduct the statistical analysis, we use 655 samples (0.1\%) from the LLaVA-655k data~\cite{liu2023improved} to generate the pruning strategy. 
To validate FitPrune, we apply it to LLaVA~\cite{liu2023visual}, LLaVA-HR~\cite{luo2024feast}, and LLaVA-NEXT~\cite{liu2024improved}. The base language models used in these MLLMs are all Vicuna ~\cite{chiang2023vicuna}. 
Under the default setting, LLaVA-1.5 7B has 576 visual tokens, while LLaVA-HR 7B stands out with 1024 visual tokens, resulting in better performance and higher computational complexity. 
For a higher resolution, LLaVA-NEXT 7B dynamically introduces visual tokens up to 2880, improving the performance by 6.0\% with 3.5$\times$ computation overhead compared to LLaVA. 
In our experiments, we randomly select 655 samples (0.1\%) from LLaVA-655k dataset~\cite{liu2023improved} to generate the pruning strategy. In our statistical process, the binary search tolerance parameter $\epsilon$ is set to 0.01. Our experiments are conducted on a single A100 40G GPU. More details can refers to our projects.

\vspace{-2mm}



\subsection{Experimental Results}


\noindent \textbf{Token redundancy analysis.} We first investigate the visual redundancy of MLLMs in Fig.\ref{fig:token_redundancy_analysis}.
From Fig.\ref{fig:token_redundancy_analysis}, the first observation is that removing a certain number of tokens after some layers of MLLMs barely affect performance. For instance, dropping about 75\% tokens at the 28-$th$ layer will not result in performance loss. 
Another observation is that the pruning strategy  greatly impacts MLLMs. In particular, pruning the tokens of the deeper layers barely impedes the performance, while pruning the shallow ones does.
For instance, pruning 50\% tokens after the 4-$th$ layer results in a 3.2\% performance drop. 
Moreover, we observe that the pruning ratios of different layers also have different results. For instance, dropping more tokens at the 16-$th$ layer will leads to more performance loss than that at the 28-$th$ or 32-$nd$ layer. 
%
Overall, these results suggest that the visual redundancy does exist in MLLMs, especially in the deeper layers. Meanwhile, the optimal pruning ratio of different layers are also obviously different. Under the manual setting, it requires numerous trials to obtain the optimal pruning recipe for MLLMs.

\noindent \textbf{Effects of FitPrune on different MLLMs.} 
%
We report the effects of FitPrune on a set of MLLMs in Tab. \ref{Tab:main} and \ref{Tab:main_additional}, including LLaVA-1.5 7B~\cite{liu2023visual}, LLaVA-HR 7B~\cite{luo2024feast} and LLaVA-NEXT 7B~\cite{liu2024improved}.  
%
%
From Tab.~\ref{Tab:main} and Tab.~\ref{Tab:main_additional}, we can first observe that our FitPrune method significantly reduces computational overhead while maintaining competitive performance of these MLLMs.
For instance, when decreasing 40\% FLOPs from visual tokens in statistic, FitPrune decreases computation overhead by 33.2\% on average without dropping performance. 
When defining a pruning ratio of 60\% FLOPs, our FitPrune reduces  the overall computation overhead of LLaVA-NEXT by 55\%, with only 0.3\% drop for VQAv2.
Similar results can be also obtained on the MLLM benchmarks, in Tab.~\ref{Tab:main_additional}.  With a 60\% pruning ratio from visual tokens, FitPrune achieves 52.4\% computation reduction, with only 0.1\% performance drop on average.
Overall, these experiments well validate the effectiveness of FitPrune in token pruning.

\begin{table}[!t]
\small
\centering
\renewcommand{\arraystretch}{1.1}
\begin{tabular}{lrc}
\toprule
\textbf{Num. of Sample} & \textbf{Search Time} & \textbf{Accuracy (\%)} \\ \midrule
\ \ \ \ \ \ \ \ \ \ \ \ 6.6K & 199 min \ \ \ \ \ \ \ \  & 61.6 \\
\ \ \ \ \ \ \ \ \ \ \ \ 2.0K  & 66 min \ \ \ \ \ \ \ \  & 61.6 \\
\ \ \ \ \ \ \ \ \ \ \ \ 0.6K  & 19 min \ \ \ \ \ \ \ \  & 61.5 \\
\ \ \ \ \ \ \ \ \ \ \ \ 65     & 2 min \ \ \ \ \ \ \ \  & 61.6 \\
\ \ \ \ \ \ \ \ \ \ \ \ 10    & 0.5 min \ \ \ \ \ \ \ \  & 61.2 \\
\bottomrule
\end{tabular}
\vspace{-1mm}
\caption{
Ablation study on the scale of data for binary search of FitPrune for LLaVA-1.5 on GQA. 
}
\vspace{-5mm}
\label{tab:ablation_results_llava_data}
\end{table}

\noindent \textbf{Comparison with SOTA pruning methods.}
We further compare our FitPrune with two representative pruning methods on LLaVA-1.5, \emph{i.e.}, ToMe~\cite{bolya2022token} and FastV~\cite{chen2024image}. 
%
From Fig.~\ref{fig:comparison}, we can observe that our FitPrune method has obvious advantages over these methods.
At lower pruning ratios, all methods maintain competitive performance, highlighting the presence of visual redundancy. For example, with a 20\% pruning ratio on the GQA dataset, ToMe experiences only a 0.2\% performance drop, and FastV shows a 0.4\% reduction, while FitPrune maintains its performance without any degradation. 
However, as the pruning ratio increases, FitPrune's advantages become more evident. With a 60\% ratio, FitPrune achieves +2.5\% and +4.5\% performance gains compared to ToMe and FastV on the VQAv2 dataset. This demonstrates FitPrune's ability to preserve model performance even under more aggressive pruning.
%
%
%
%
With a larger reduction target, FitPrune can better maintain the performance.
For instance, when reducing FLOPs by 60\%, FitPrune only drops 0.9\% performance on VQAv2, whereas FastV loses 4.5\%.
In summary, compared with other methods, the proposed FitPrune method shows robustness as the pruning ratio increases, proving its effectiveness in the token pruning task.

\noindent \textbf{Ablation study.} 
Next, we ablate the key settings of FitPrune. We first compare the performance of different pruning ratios, and also visualize the attention distributions after pruning in Fig.~\ref{fig:abla}-(a) and (b).
From these figures, we can first see that FitPrune is robust and can well maintain the high performance when increasing the pruning ratio. Besides, we can also see that the fit of distributions also affect the final performance. For instance, when the pruning ratio is 80\%, the average divergence of the self-attention and cross-attention distributions compared to the original distributions reaches approximately 27\%, leading to a performance drop of 3.6\%. Furthermore, the divergence caused by token pruning is relatively uniform across layers. However, compared to cross-attention, self-attention is more unstable and significantly affected by token pruning, indicating that MLLMs rely more heavily on text tokens to gather information from visual tokens.
Overall, these results well validate our assumptions about the distribution fitting and the consideration of both self- and cross-attention modeling.
%
Additionally, we also visualize the pruning strategy in Fig.~\ref{fig:abla}-(c). 
It can be seen that as the pruning ratios increase, FitPrune tends to focus more on pruning tokens in the shallower layers, leading to a significant reduction in computational overhead. 

\begin{figure}[!t]
\centering
\includegraphics[width=1.0\linewidth]{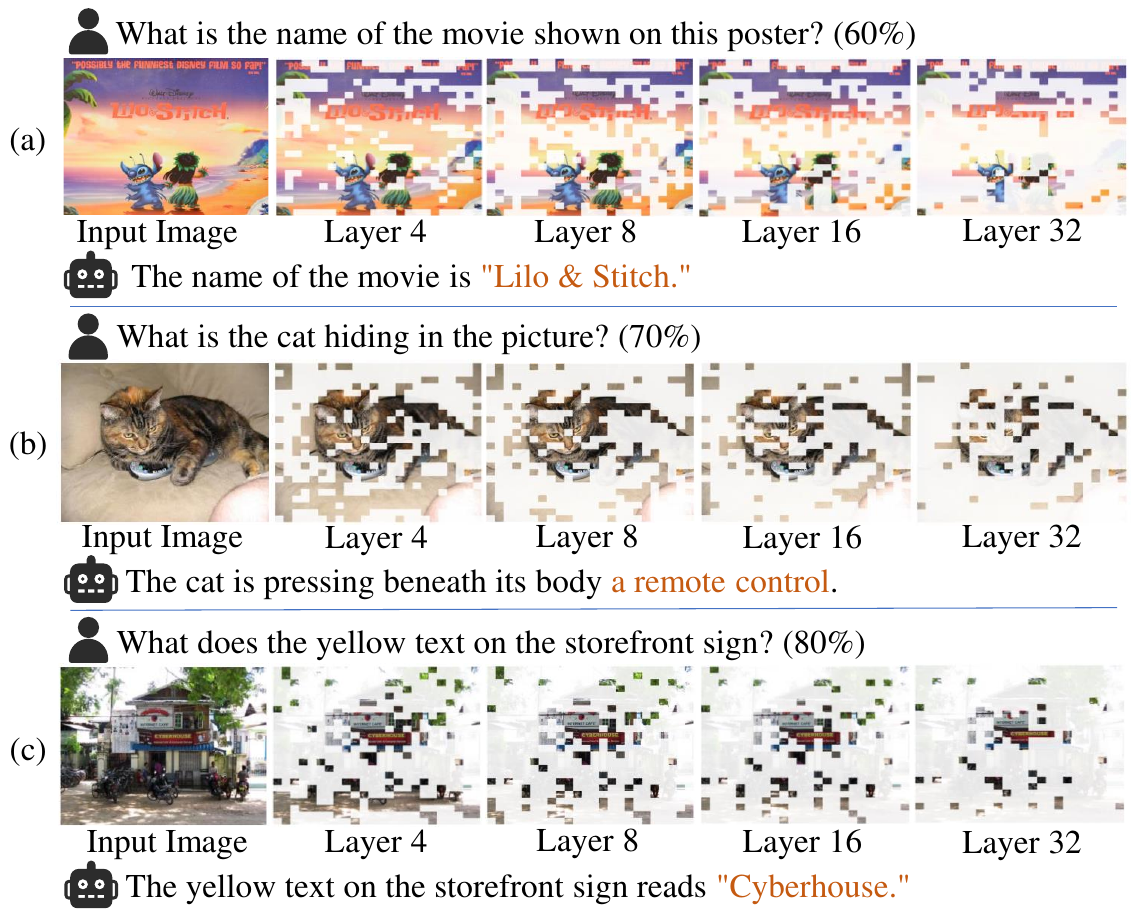}
\vspace{-0.4cm}
\caption{
Visualization of token pruning process.
The percentage in brackets indicates the pruning ratio.
}
\vspace{-4mm}
\label{fig:qualitative}
\end{figure}

In Tab.~\ref{tab:ablation_results_llava_data}, we ablate the number of examples for FitPrune to produce a pruning recipe.
The first observation is that our FitPrune method maintains stable performance even with a small scale of statistical data.
For instance, when using only 0.01\% of the dataset (65 samples), the model achieves an accuracy of 61.6\%, comparable to using 1\% of the dataset (6.6K samples). 
This result indicates that the pattern of information exchange may depend more on the nature of the model itself, and our FitPrune method can capture this pattern efficiently.
Furthermore, as the amount of data decreases, the statistical time is also reduced substantially. 
Overall, these experiment results well confirm the motivation and efficiency of FitPrune.

\noindent \textbf{Visualization results.} 
In Fig.~\ref{fig:qualitative}, we visualize the token pruning process for 3 challenging samples on LLaVA-1.5 7B. 
From Fig.~\ref{fig:qualitative}, it can be seen that our FitPrune retains key object tokens according to the question while pruning irrelevant tokens. 
For example, our FitPrune retains the poster title in Fig.~\ref{fig:qualitative}-(a).
Besides, as the number of deleted tokens increases, tokens from the primary object are gradually pruned.
In the process, we can find that the tokens with the most complex texture are still retained.
As shown in Fig.~\ref{fig:qualitative}-(b), the head of the cat is retained to the end.
For the text-related tasks, we can notice that FitPrune retains all text-related tokens.
For instance, in Fig.~\ref{fig:qualitative}-(c), the yellow text is preserved intact until the end. 
These visualizations highlight FitPrune's effectiveness in retaining key tokens essential for accurate question answering.


\vspace{-2mm}

\section{Conclusion}

In this paper, we introduce FitPrune, a novel and training-free approach for visual token pruning  of MLLMs. By framing token pruning as a statistical problem, FitPrune aims to minimize divergence in attention distributions, enabling the efficient pruning of redundant tokens.  FitPrune can generate an optimal pruning strategy based on a small batch of data, avoiding costly  manual trials. We validate FitPrune on various MLLMs, demonstrating its ability to reduce computational overhead while preserving performance. FitPrune offers a practical and efficient solution for optimizing MLLMs, and can be seamlessly integrated into existing models, paving the way for more scalable and efficient vision-language tasks.

\section{Acknowledgments}
This work was supported by the National Science Fund for Distinguished Young Scholars (No.62025603), the National Natural Science Foundation of China (No. U21B2037, No. U22B2051, No. U23A20383, No. U21A20472, No. 62176222, No. 62176223, No. 62176226, No. 62072386, No. 62072387, No. 62072389, No. 62002305 and No. 62272401), and the Natural Science Foundation of Fujian Province of China (No. 2021J06003, No.2022J06001).

\bibliography{aaai25}

\begin{thebibliography}{55}
\providecommand{\natexlab}[1]{#1}

\bibitem[{Anagnostidis et~al.(2024)Anagnostidis, Pavllo, Biggio, Noci, Lucchi, and Hofmann}]{anagnostidis2024dynamic}
Anagnostidis, S.; Pavllo, D.; Biggio, L.; Noci, L.; Lucchi, A.; and Hofmann, T. 2024.
\newblock Dynamic context pruning for efficient and interpretable autoregressive transformers.
\newblock \emph{Advances in Neural Information Processing Systems}, 36.

\bibitem[{Bai et~al.(2023)Bai, Bai, Yang, Wang, Tan, Wang, Lin, Zhou, and Zhou}]{bai2023qwen}
Bai, J.; Bai, S.; Yang, S.; Wang, S.; Tan, S.; Wang, P.; Lin, J.; Zhou, C.; and Zhou, J. 2023.
\newblock Qwen-vl: A frontier large vision-language model with versatile abilities.
\newblock \emph{arXiv preprint arXiv:2308.12966}.

\bibitem[{Bao et~al.(2022)Bao, Wang, Dong, Liu, Mohammed, Aggarwal, Som, Piao, and Wei}]{bao2022vlmo}
Bao, H.; Wang, W.; Dong, L.; Liu, Q.; Mohammed, O.~K.; Aggarwal, K.; Som, S.; Piao, S.; and Wei, F. 2022.
\newblock Vlmo: Unified vision-language pre-training with mixture-of-modality-experts.
\newblock \emph{Adv. Neural Inform. Process. Syst.}, 35: 32897--32912.

\bibitem[{Bolya et~al.(2022)Bolya, Fu, Dai, Zhang, Feichtenhofer, and Hoffman}]{bolya2022token}
Bolya, D.; Fu, C.-Y.; Dai, X.; Zhang, P.; Feichtenhofer, C.; and Hoffman, J. 2022.
\newblock Token merging: Your vit but faster.
\newblock \emph{arXiv preprint arXiv:2210.09461}.

\bibitem[{Cao et~al.(2024)Cao, Ye, Li, Yu, Tang, Lu, and Chen}]{cao2024madtp}
Cao, J.; Ye, P.; Li, S.; Yu, C.; Tang, Y.; Lu, J.; and Chen, T. 2024.
\newblock MADTP: Multimodal Alignment-Guided Dynamic Token Pruning for Accelerating Vision-Language Transformer.
\newblock \emph{arXiv preprint arXiv:2403.02991}.

\bibitem[{Chen et~al.(2024{\natexlab{a}})Chen, Zhao, Liu, Bai, Lin, Zhou, and Chang}]{chen2024image}
Chen, L.; Zhao, H.; Liu, T.; Bai, S.; Lin, J.; Zhou, C.; and Chang, B. 2024{\natexlab{a}}.
\newblock An Image is Worth 1/2 Tokens After Layer 2: Plug-and-Play Inference Acceleration for Large Vision-Language Models.
\newblock \emph{arXiv preprint arXiv:2403.06764}.

\bibitem[{Chen et~al.(2023)Chen, Shao, Xu, Lin, Zhang, Chao, Ji, Qiao, and Luo}]{chen2023diffrate}
Chen, M.; Shao, W.; Xu, P.; Lin, M.; Zhang, K.; Chao, F.; Ji, R.; Qiao, Y.; and Luo, P. 2023.
\newblock Diffrate: Differentiable compression rate for efficient vision transformers.
\newblock In \emph{Proceedings of the IEEE/CVF International Conference on Computer Vision}, 17164--17174.

\bibitem[{Chen et~al.(2024{\natexlab{b}})Chen, Wu, Wang, Su, Chen, Xing, Zhong, Zhang, Zhu, Lu et~al.}]{chen2024internvl}
Chen, Z.; Wu, J.; Wang, W.; Su, W.; Chen, G.; Xing, S.; Zhong, M.; Zhang, Q.; Zhu, X.; Lu, L.; et~al. 2024{\natexlab{b}}.
\newblock Internvl: Scaling up vision foundation models and aligning for generic visual-linguistic tasks.
\newblock In \emph{Proceedings of the IEEE/CVF Conference on Computer Vision and Pattern Recognition}, 24185--24198.

\bibitem[{Chiang et~al.(2023)Chiang, Li, Lin, Sheng, Wu, Zhang, Zheng, Zhuang, Zhuang, Gonzalez et~al.}]{chiang2023vicuna}
Chiang, W.-L.; Li, Z.; Lin, Z.; Sheng, Y.; Wu, Z.; Zhang, H.; Zheng, L.; Zhuang, S.; Zhuang, Y.; Gonzalez, J.~E.; et~al. 2023.
\newblock Vicuna: An open-source chatbot impressing gpt-4 with 90\%* chatgpt quality.
\newblock \emph{See https://vicuna. lmsys. org (accessed 14 April 2023)}.

\bibitem[{Chu et~al.(2024)Chu, Qiao, Zhang, Xu, Wei, Yang, Sun, Hu, Lin, Zhang et~al.}]{chu2024mobilevlm}
Chu, X.; Qiao, L.; Zhang, X.; Xu, S.; Wei, F.; Yang, Y.; Sun, X.; Hu, Y.; Lin, X.; Zhang, B.; et~al. 2024.
\newblock Mobilevlm v2: Faster and stronger baseline for vision language model.
\newblock \emph{arXiv preprint arXiv:2402.03766}.

\bibitem[{Dao et~al.(2022)Dao, Fu, Ermon, Rudra, and R{\'e}}]{dao2022flashattention}
Dao, T.; Fu, D.; Ermon, S.; Rudra, A.; and R{\'e}, C. 2022.
\newblock Flashattention: Fast and memory-efficient exact attention with io-awareness.
\newblock \emph{Advances in Neural Information Processing Systems}, 35: 16344--16359.

\bibitem[{Dong et~al.(2023)Dong, Sun, Lu, Xie, Liu, Kong, Meng, Li, Lin, Fang et~al.}]{dong2023heatvit}
Dong, P.; Sun, M.; Lu, A.; Xie, Y.; Liu, K.; Kong, Z.; Meng, X.; Li, Z.; Lin, X.; Fang, Z.; et~al. 2023.
\newblock Heatvit: Hardware-efficient adaptive token pruning for vision transformers.
\newblock In \emph{2023 IEEE International Symposium on High-Performance Computer Architecture (HPCA)}, 442--455. IEEE.

\bibitem[{Dosovitskiy et~al.(2020)Dosovitskiy, Beyer, Kolesnikov, Weissenborn, Zhai, Unterthiner, Dehghani, Minderer, Heigold, Gelly et~al.}]{dosovitskiy2020image}
Dosovitskiy, A.; Beyer, L.; Kolesnikov, A.; Weissenborn, D.; Zhai, X.; Unterthiner, T.; Dehghani, M.; Minderer, M.; Heigold, G.; Gelly, S.; et~al. 2020.
\newblock An image is worth 16x16 words: Transformers for image recognition at scale.
\newblock \emph{arXiv preprint arXiv:2010.11929}.

\bibitem[{Fayyaz et~al.(2022)Fayyaz, Koohpayegani, Jafari, Sengupta, Joze, Sommerlade, Pirsiavash, and Gall}]{fayyaz2022adaptive}
Fayyaz, M.; Koohpayegani, S.~A.; Jafari, F.~R.; Sengupta, S.; Joze, H. R.~V.; Sommerlade, E.; Pirsiavash, H.; and Gall, J. 2022.
\newblock Adaptive token sampling for efficient vision transformers.
\newblock In \emph{European Conference on Computer Vision}, 396--414. Springer.

\bibitem[{Fu et~al.(2023)Fu, Chen, Shen, Qin, Zhang, Lin, Qiu, Lin, Yang, Zheng et~al.}]{fu2023mme}
Fu, C.; Chen, P.; Shen, Y.; Qin, Y.; Zhang, M.; Lin, X.; Qiu, Z.; Lin, W.; Yang, J.; Zheng, X.; et~al. 2023.
\newblock MME: A Comprehensive Evaluation Benchmark for Multimodal Large Language Models.
\newblock \emph{arXiv preprint arXiv:2306.13394}.

\bibitem[{Goyal et~al.(2017)Goyal, Khot, Summers{-}Stay, Batra, and Parikh}]{conf/cvpr/GoyalKSBP17}
Goyal, Y.; Khot, T.; Summers{-}Stay, D.; Batra, D.; and Parikh, D. 2017.
\newblock Making the {V} in {VQA} Matter: Elevating the Role of Image Understanding in Visual Question Answering.
\newblock In \emph{Proceedings of the IEEE/CVF Conference on Computer Vision and Pattern Recognition (CVPR)}, 6325--6334.

\bibitem[{Gurari et~al.(2018)Gurari, Li, Stangl, Guo, Lin, Grauman, Luo, and Bigham}]{gurari2018vizwiz}
Gurari, D.; Li, Q.; Stangl, A.~J.; Guo, A.; Lin, C.; Grauman, K.; Luo, J.; and Bigham, J.~P. 2018.
\newblock Vizwiz grand challenge: Answering visual questions from blind people.
\newblock In \emph{Proceedings of the IEEE conference on computer vision and pattern recognition}, 3608--3617.

\bibitem[{Han et~al.(2021)Han, Huang, Song, Yang, Wang, and Wang}]{han2021dynamic}
Han, Y.; Huang, G.; Song, S.; Yang, L.; Wang, H.; and Wang, Y. 2021.
\newblock Dynamic neural networks: A survey.
\newblock \emph{IEEE Transactions on Pattern Analysis and Machine Intelligence}, 44(11): 7436--7456.

\bibitem[{Huang et~al.(2024)Huang, Liu, Guo, and Gong}]{huang2024visual}
Huang, W.; Liu, H.; Guo, M.; and Gong, N.~Z. 2024.
\newblock Visual hallucinations of multi-modal large language models.
\newblock \emph{arXiv preprint arXiv:2402.14683}.

\bibitem[{Hudson and Manning(2019)}]{hudson2019gqa}
Hudson, D.~A.; and Manning, C.~D. 2019.
\newblock Gqa: A new dataset for real-world visual reasoning and compositional question answering.
\newblock In \emph{CVPR}, 6700--6709.

\bibitem[{Kenton and Toutanova(2019)}]{devlin2018bert}
Kenton, J. D. M.-W.~C.; and Toutanova, L.~K. 2019.
\newblock BERT: Pre-training of Deep Bidirectional Transformers for Language Understanding.
\newblock In \emph{NAACL-HLT}, 4171--4186.

\bibitem[{Kim et~al.(2022)Kim, Shen, Thorsley, Gholami, Kwon, Hassoun, and Keutzer}]{kim2022learned}
Kim, S.; Shen, S.; Thorsley, D.; Gholami, A.; Kwon, W.; Hassoun, J.; and Keutzer, K. 2022.
\newblock Learned token pruning for transformers.
\newblock In \emph{Proceedings of the 28th ACM SIGKDD Conference on Knowledge Discovery and Data Mining}, 784--794.

\bibitem[{Kong et~al.(2022)Kong, Dong, Ma, Meng, Niu, Sun, Shen, Yuan, Ren, Tang et~al.}]{kong2022spvit}
Kong, Z.; Dong, P.; Ma, X.; Meng, X.; Niu, W.; Sun, M.; Shen, X.; Yuan, G.; Ren, B.; Tang, H.; et~al. 2022.
\newblock Spvit: Enabling faster vision transformers via latency-aware soft token pruning.
\newblock In \emph{European conference on computer vision}, 620--640. Springer.

\bibitem[{Le~Scao et~al.(2023)Le~Scao, Fan, Akiki, Pavlick, Ili{\'c}, Hesslow, Castagn{\'e}, Luccioni, Yvon, Gall{\'e} et~al.}]{le2023bloom}
Le~Scao, T.; Fan, A.; Akiki, C.; Pavlick, E.; Ili{\'c}, S.; Hesslow, D.; Castagn{\'e}, R.; Luccioni, A.~S.; Yvon, F.; Gall{\'e}, M.; et~al. 2023.
\newblock Bloom: A 176b-parameter open-access multilingual language model.

\bibitem[{Li et~al.(2023{\natexlab{a}})Li, Li, Savarese, and Hoi}]{li2023blip}
Li, J.; Li, D.; Savarese, S.; and Hoi, S. 2023{\natexlab{a}}.
\newblock BLIP-2: Bootstrapping Language-Image Pre-training with Frozen Image Encoders and Large Language Models.
\newblock In \emph{International Conference on Machine Learning}.

\bibitem[{Li et~al.(2023{\natexlab{b}})Li, Du, Zhou, Wang, Zhao, and Wen}]{li2023evaluatingpope}
Li, Y.; Du, Y.; Zhou, K.; Wang, J.; Zhao, W.~X.; and Wen, J.-R. 2023{\natexlab{b}}.
\newblock Evaluating object hallucination in large vision-language models.
\newblock \emph{arXiv preprint arXiv:2305.10355}.

\bibitem[{Li et~al.(2023{\natexlab{c}})Li, Yang, Liu, Ma, Zhang, Yang, Sun, Liu, and Bai}]{li2023monkey}
Li, Z.; Yang, B.; Liu, Q.; Ma, Z.; Zhang, S.; Yang, J.; Sun, Y.; Liu, Y.; and Bai, X. 2023{\natexlab{c}}.
\newblock Monkey: Image resolution and text label are important things for large multi-modal models.
\newblock \emph{arXiv preprint arXiv:2311.06607}.

\bibitem[{Lin et~al.(2023)Lin, Zhu, Ye, Ning, Jin, and Yuan}]{lin2023video}
Lin, B.; Zhu, B.; Ye, Y.; Ning, M.; Jin, P.; and Yuan, L. 2023.
\newblock Video-llava: Learning united visual representation by alignment before projection.
\newblock \emph{arXiv preprint arXiv:2311.10122}.

\bibitem[{Liu et~al.(2023{\natexlab{a}})Liu, Tao, Liang, Feng, Shen, Huang, and Zhao}]{liu2023length}
Liu, C.; Tao, C.; Liang, J.; Feng, J.; Shen, T.; Huang, Q.; and Zhao, D. 2023{\natexlab{a}}.
\newblock Length-Adaptive Distillation: Customizing Small Language Model for Dynamic Token Pruning.
\newblock In \emph{The 2023 Conference on Empirical Methods in Natural Language Processing}.

\bibitem[{Liu et~al.(2023{\natexlab{b}})Liu, Li, Li, and Lee}]{liu2023improved}
Liu, H.; Li, C.; Li, Y.; and Lee, Y.~J. 2023{\natexlab{b}}.
\newblock Improved baselines with visual instruction tuning.
\newblock \emph{arXiv preprint arXiv:2310.03744}.

\bibitem[{Liu et~al.(2024)Liu, Li, Li, and Lee}]{liu2024improved}
Liu, H.; Li, C.; Li, Y.; and Lee, Y.~J. 2024.
\newblock Improved baselines with visual instruction tuning.
\newblock In \emph{Proceedings of the IEEE/CVF Conference on Computer Vision and Pattern Recognition}, 26296--26306.

\bibitem[{Liu et~al.(2023{\natexlab{c}})Liu, Li, Wu, and Lee}]{liu2023visual}
Liu, H.; Li, C.; Wu, Q.; and Lee, Y.~J. 2023{\natexlab{c}}.
\newblock Visual instruction tuning.
\newblock \emph{arXiv preprint arXiv:2304.08485}.

\bibitem[{Liu et~al.(2023{\natexlab{d}})Liu, Duan, Zhang, Li, Zhang, Zhao, Yuan, Wang, He, Liu et~al.}]{liu2023mmbench}
Liu, Y.; Duan, H.; Zhang, Y.; Li, B.; Zhang, S.; Zhao, W.; Yuan, Y.; Wang, J.; He, C.; Liu, Z.; et~al. 2023{\natexlab{d}}.
\newblock Mmbench: Is your multi-modal model an all-around player?
\newblock \emph{arXiv preprint arXiv:2307.06281}.

\bibitem[{Liu et~al.(2023{\natexlab{e}})Liu, Li, Yang, Li, Yin, Liu, Jin, and Bai}]{liu2023hidden}
Liu, Y.; Li, Z.; Yang, B.; Li, C.; Yin, X.; Liu, C.-l.; Jin, L.; and Bai, X. 2023{\natexlab{e}}.
\newblock On the hidden mystery of ocr in large multimodal models.
\newblock \emph{arXiv preprint arXiv:2305.07895}.

\bibitem[{Lu et~al.(2022)Lu, Mishra, Xia, Qiu, Chang, Zhu, Tafjord, Clark, and Kalyan}]{lu2022learn}
Lu, P.; Mishra, S.; Xia, T.; Qiu, L.; Chang, K.-W.; Zhu, S.-C.; Tafjord, O.; Clark, P.; and Kalyan, A. 2022.
\newblock Learn to explain: Multimodal reasoning via thought chains for science question answering.
\newblock \emph{Advances in Neural Information Processing Systems}, 35: 2507--2521.

\bibitem[{Luo et~al.(2024)Luo, Zhou, Zhang, Zheng, Sun, and Ji}]{luo2024feast}
Luo, G.; Zhou, Y.; Zhang, Y.; Zheng, X.; Sun, X.; and Ji, R. 2024.
\newblock Feast Your Eyes: Mixture-of-Resolution Adaptation for Multimodal Large Language Models.
\newblock \emph{arXiv preprint arXiv:2403.03003}.

\bibitem[{Meng et~al.(2022)Meng, Li, Chen, Lan, Wu, Jiang, and Lim}]{meng2022adavit}
Meng, L.; Li, H.; Chen, B.-C.; Lan, S.; Wu, Z.; Jiang, Y.-G.; and Lim, S.-N. 2022.
\newblock Adavit: Adaptive vision transformers for efficient image recognition.
\newblock In \emph{Proceedings of the IEEE/CVF Conference on Computer Vision and Pattern Recognition}, 12309--12318.

\bibitem[{OpenAI(2023)}]{openai2023gpt}
OpenAI, R. 2023.
\newblock GPT-4 technical report.
\newblock \emph{arXiv preprint arXiv:2303.08774}.

\bibitem[{Radford et~al.(2021)Radford, Kim, Hallacy, Ramesh, Goh, Agarwal, Sastry, Askell, Mishkin, Clark et~al.}]{radford2021learning}
Radford, A.; Kim, J.~W.; Hallacy, C.; Ramesh, A.; Goh, G.; Agarwal, S.; Sastry, G.; Askell, A.; Mishkin, P.; Clark, J.; et~al. 2021.
\newblock Learning transferable visual models from natural language supervision.
\newblock In \emph{International conference on machine learning}, 8748--8763. PMLR.

\bibitem[{Rao et~al.(2021)Rao, Zhao, Liu, Lu, Zhou, and Hsieh}]{rao2021dynamicvit}
Rao, Y.; Zhao, W.; Liu, B.; Lu, J.; Zhou, J.; and Hsieh, C.-J. 2021.
\newblock Dynamicvit: Efficient vision transformers with dynamic token sparsification.
\newblock \emph{Advances in neural information processing systems}, 34: 13937--13949.

\bibitem[{Singh et~al.(2019)Singh, Natarajan, Shah, Jiang, Chen, Batra, Parikh, and Rohrbach}]{singh2019towards}
Singh, A.; Natarajan, V.; Shah, M.; Jiang, Y.; Chen, X.; Batra, D.; Parikh, D.; and Rohrbach, M. 2019.
\newblock Towards vqa models that can read.
\newblock In \emph{Proceedings of the IEEE/CVF conference on computer vision and pattern recognition}, 8317--8326.

\bibitem[{Tong et~al.(2024)Tong, Liu, Zhai, Ma, LeCun, and Xie}]{tong2024eyes}
Tong, S.; Liu, Z.; Zhai, Y.; Ma, Y.; LeCun, Y.; and Xie, S. 2024.
\newblock Eyes wide shut? exploring the visual shortcomings of multimodal llms.
\newblock In \emph{Proceedings of the IEEE/CVF Conference on Computer Vision and Pattern Recognition}, 9568--9578.

\bibitem[{Touvron et~al.(2023)Touvron, Lavril, Izacard, Martinet, Lachaux, Lacroix, Rozi{\`e}re, Goyal, Hambro, Azhar et~al.}]{touvron2023llama}
Touvron, H.; Lavril, T.; Izacard, G.; Martinet, X.; Lachaux, M.-A.; Lacroix, T.; Rozi{\`e}re, B.; Goyal, N.; Hambro, E.; Azhar, F.; et~al. 2023.
\newblock Llama: Open and efficient foundation language models.
\newblock \emph{arXiv preprint arXiv:2302.13971}.

\bibitem[{Vaswani et~al.(2017)Vaswani, Shazeer, Parmar, Uszkoreit, Jones, Gomez, Kaiser, and Polosukhin}]{vaswani2017attention}
Vaswani, A.; Shazeer, N.; Parmar, N.; Uszkoreit, J.; Jones, L.; Gomez, A.~N.; Kaiser, {\L}.; and Polosukhin, I. 2017.
\newblock Attention is all you need.
\newblock \emph{Advances in neural information processing systems}, 30.

\bibitem[{Wang, Dedhia, and Jha(2023)}]{wang2023zero}
Wang, H.; Dedhia, B.; and Jha, N.~K. 2023.
\newblock Zero-TPrune: Zero-Shot Token Pruning through Leveraging of the Attention Graph in Pre-Trained Transformers.
\newblock \emph{arXiv preprint arXiv:2305.17328}.

\bibitem[{Wang et~al.(2024)Wang, Bai, Tan, Wang, Fan, Bai, Chen, Liu, Wang, Ge et~al.}]{wang2024qwen2}
Wang, P.; Bai, S.; Tan, S.; Wang, S.; Fan, Z.; Bai, J.; Chen, K.; Liu, X.; Wang, J.; Ge, W.; et~al. 2024.
\newblock Qwen2-vl: Enhancing vision-language model's perception of the world at any resolution.
\newblock \emph{arXiv preprint arXiv:2409.12191}.

\bibitem[{Wang et~al.(2023)Wang, Lv, Yu, Hong, Qi, Wang, Ji, Yang, Zhao, Song et~al.}]{wang2023cogvlm}
Wang, W.; Lv, Q.; Yu, W.; Hong, W.; Qi, J.; Wang, Y.; Ji, J.; Yang, Z.; Zhao, L.; Song, X.; et~al. 2023.
\newblock Cogvlm: Visual expert for pretrained language models.
\newblock \emph{arXiv preprint arXiv:2311.03079}.

\bibitem[{Wei et~al.(2023)Wei, Ye, Zhang, Tang, and Liang}]{wei2023joint}
Wei, S.; Ye, T.; Zhang, S.; Tang, Y.; and Liang, J. 2023.
\newblock Joint token pruning and squeezing towards more aggressive compression of vision transformers.
\newblock In \emph{Proceedings of the IEEE/CVF Conference on Computer Vision and Pattern Recognition}, 2092--2101.

\bibitem[{Xu et~al.(2023)Xu, Ye, Yan, Shi, Ye, Xu, Li, Bi, Qian, Wang et~al.}]{xu2023mplug}
Xu, H.; Ye, Q.; Yan, M.; Shi, Y.; Ye, J.; Xu, Y.; Li, C.; Bi, B.; Qian, Q.; Wang, W.; et~al. 2023.
\newblock mplug-2: A modularized multi-modal foundation model across text, image and video.
\newblock In \emph{International Conference on Machine Learning}, 38728--38748. PMLR.

\bibitem[{Ye et~al.(2021)Ye, Lin, Huang, and Sun}]{ye2021tr}
Ye, D.; Lin, Y.; Huang, Y.; and Sun, M. 2021.
\newblock Tr-bert: Dynamic token reduction for accelerating bert inference.
\newblock \emph{arXiv preprint arXiv:2105.11618}.

\bibitem[{Yu et~al.(2023)Yu, Yang, Li, Wang, Lin, Liu, Wang, and Wang}]{yu2023mmvet}
Yu, W.; Yang, Z.; Li, L.; Wang, J.; Lin, K.; Liu, Z.; Wang, X.; and Wang, L. 2023.
\newblock Mm-vet: Evaluating large multimodal models for integrated capabilities.
\newblock \emph{arXiv preprint arXiv:2308.02490}.

\bibitem[{Zhang et~al.(2022)Zhang, Roller, Goyal, Artetxe, Chen, Chen, Dewan, Diab, Li, Lin et~al.}]{zhang2205opt}
Zhang, S.; Roller, S.; Goyal, N.; Artetxe, M.; Chen, M.; Chen, S.; Dewan, C.; Diab, M.; Li, X.; Lin, X.~V.; et~al. 2022.
\newblock Opt: Open pre-trained transformer language models.
\newblock \emph{URL https://arxiv. org/abs/2205.01068}.

\bibitem[{Zhou et~al.(2020)Zhou, Ji, Sun, Luo, Hong, Su, Ding, and Shao}]{zhou2020k}
Zhou, Y.; Ji, R.; Sun, X.; Luo, G.; Hong, X.; Su, J.; Ding, X.; and Shao, L. 2020.
\newblock K-armed bandit based multi-modal network architecture search for visual question answering.
\newblock In \emph{Proceedings of the 28th ACM international conference on multimedia}, 1245--1254.

\bibitem[{Zhou et~al.(2021)Zhou, Ren, Zhu, Sun, Liu, Ding, Xu, and Ji}]{zhou2021trar}
Zhou, Y.; Ren, T.; Zhu, C.; Sun, X.; Liu, J.; Ding, X.; Xu, M.; and Ji, R. 2021.
\newblock Trar: Routing the attention spans in transformer for visual question answering.
\newblock In \emph{Proceedings of the IEEE/CVF international conference on computer vision}, 2074--2084.

\bibitem[{Zhu et~al.(2023)Zhu, Chen, Shen, Li, and Elhoseiny}]{zhu2023minigpt}
Zhu, D.; Chen, J.; Shen, X.; Li, X.; and Elhoseiny, M. 2023.
\newblock Minigpt-4: Enhancing vision-language understanding with advanced large language models.
\newblock \emph{arXiv preprint arXiv:2304.10592}.

\end{thebibliography}

\appendix
\section{Appendix}
\label{sec:Appendix}

\begin{figure}[!ht]
\centering
\includegraphics[width=0.9\columnwidth]{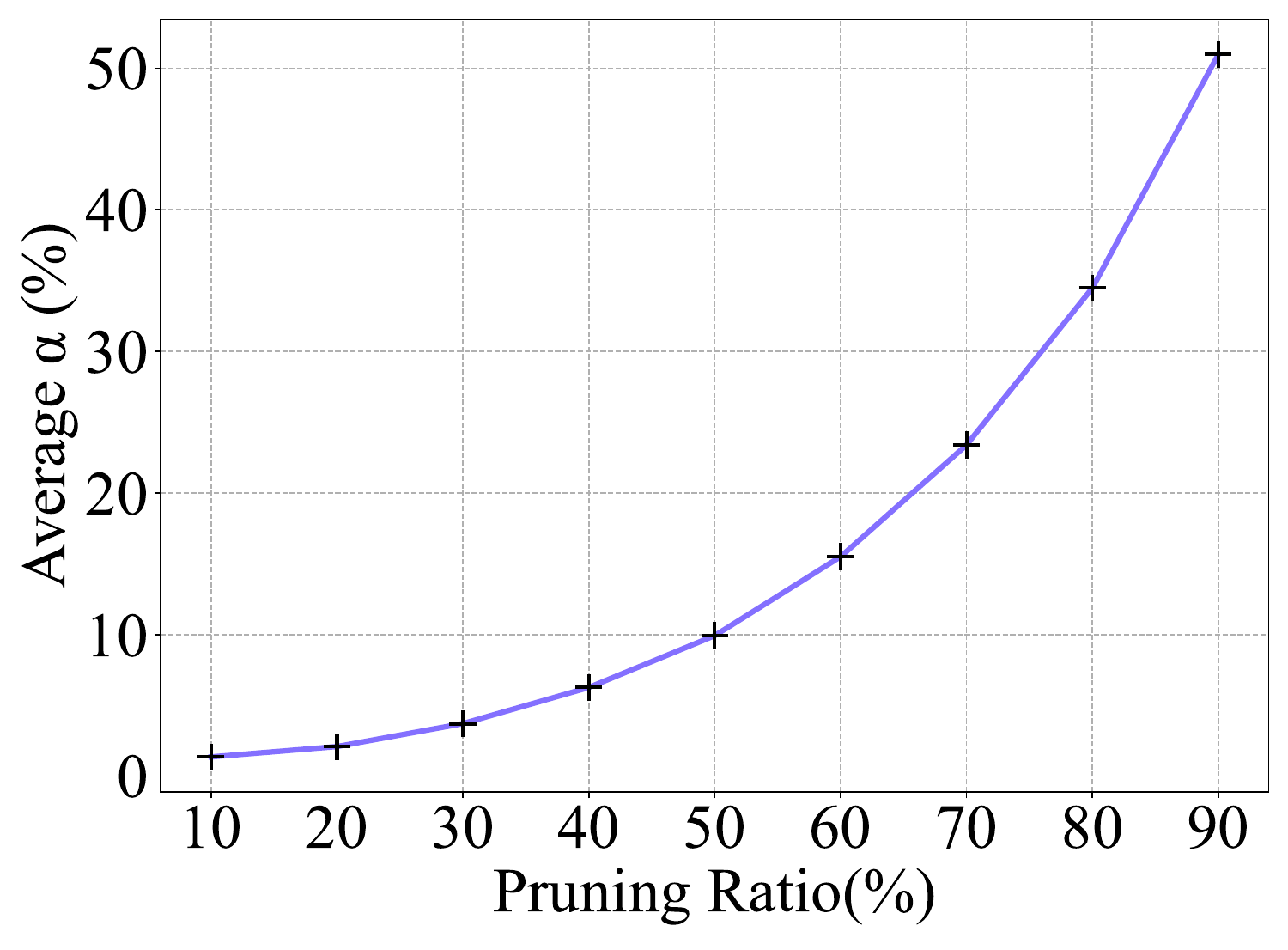}
\vspace{-3mm}
\caption{
Relationship between pruning ratio and average divergence $\alpha$ used in the statistical analysis.
}
\label{fig:p_alpha}
\end{figure}
\vspace{-3mm}

\begin{table}[!h]
\centering
\renewcommand{\arraystretch}{1.1}
\setlength{\tabcolsep}{1.0mm}
\centering

\begin{tabular}{cccc}
\toprule
\makecell[c]{\textbf{Pruning} \\ \textbf{Ratio (\%)}}
& \textbf{Accuracy (\%)} 
& \textbf{TFLOPs} 
& \makecell[c]{\textbf{Throughput} \\ \textbf{(samples/s)}}\\
\midrule
\rowcolor{gray!20} 0 & 62.0 & 9.1 & 6.8 \\
10 & 62.0 & 8.6 ($- 5.6\%$) & 7.0 ($+ 4.1\%$)\\
20 & 62.0 & 7.9 ($- 13.3\%$) & 7.4 ($+ 9.1\%$)\\
30 & 62.0 & 6.9 ($- 24.3\%$) & 7.7 ($+ 14.0\%$)\\
40 & 61.9 & 6.0 ($- 33.7\%$) & 8.0 ($+ 17.8\%$)\\
50 & 61.9 & 5.2 ($- 43.0\%$) & 8.3 ($+ 22.2\%$)\\
60 & 61.5 & 4.4 ($- 51.6\%$) &  8.6 ($+ 26.8\%$)\\
70 & 60.3 & 3.6 ($- 60.2\%$) &  9.0 ($+ 33.5\%$)\\
80 & 58.4 & 2.8 ($- 68.7\%$) & 9.5 ($+ 40.2\%$)\\
90 & 52.4 & 2.1 ($- 77.2\%$) & 10.0 ($+ 47.5\%$)\\
\bottomrule
\end{tabular}
\caption{
Ablation experiment results on the GQA dataset under different pruning ratios.
}
\label{tab:ablation_results}
\vspace{-5mm}
\end{table}

\begin{table}[!t]
\centering
\renewcommand{\arraystretch}{1.1} 
\setlength{\tabcolsep}{8pt} 
\begin{tabular}{lccc}
\toprule
\textbf{Method}        & \textbf{TextVQA} & \textbf{GQA} & \textbf{MME} \\ \midrule
\rowcolor{gray!20} LLaVA1.5 7B            & 58.2            & 62.0         & 1510.7       \\
Cross                  & 57.6            & 60.1         & 1491.0       \\
Self                   & 51.6            & 55.4         & 1263.3       \\
Self + Cross           & \textbf{58.2}   & \textbf{61.5}& \textbf{1507.9} \\ 
\bottomrule
\end{tabular}
\caption{
Ablation study of attention distributions under a pruning ratio of 60\% on TextVQA, GQA, and MME datasets. 
}
\label{tab:ablation_attention_distributions}
\end{table}

\begin{table}[!t]
\centering
\setlength{\tabcolsep}{0.8pt} 
\begin{tabular}{lcccc}
\toprule
\textbf{Model} & \textbf{TextVQA} & \textbf{MMBench} & \textbf{MME} & \textbf{TFLOPs} \\ 
\midrule
\rowcolor{gray!20} MobileVLMv2 3B         & 57.5 & 63.2 & 1440.5 & 1.2 \\
+FitPrune 60\%         & 57.0 & 61.8 & 1414.4 & 0.7 \\
\rowcolor{gray!20} LLaVA-1.5 7B           & 58.2 & 64.3 & 1510.7 & 8.8 \\
+FitPrune 60\%         & 58.2 & 64.6 & 1507.9 & 4.2 \\
\rowcolor{gray!20} LLaVA-1.5 13B          & 61.3 & 67.7 & 1531.3 & 17.2 \\
+FitPrune 60\%         & 60.9 & 68.2 & 1519.8 & 8.2 \\
\rowcolor{gray!20} QwenVL2 7B             & 84.3 & 77.8 & 1673.1 & 25.5 \\
+FitPrune 60\%         & 83.6 & 78.3 & 1642.0 & 14.4 \\
\rowcolor{gray!20} InternVL2 8B           & 77.6 & 82.4 & 1647.3 & 26.8 \\
+FitPrune 60\%         & 76.5 & 81.8 & 1616.3 & 13.3 \\
\bottomrule
\end{tabular}
\caption{
Generalization results of FitPrune on various MLLMs with different scales and families. "+FitPrune 60\%" indicates results with a 60\% pruning ratio.
}
\label{tab:fitprune_generalization}
\end{table}

\begin{table*}[t]
\centering
\setlength{\tabcolsep}{4pt} 
\begin{tabular}{lcccccc}
\toprule
\textbf{Method} & \textbf{TextVQA} & \textbf{Samples/sec} & \textbf{SQA} & \textbf{Samples/sec} & \textbf{GQA} & \textbf{Samples/sec} \\ 
\midrule
\rowcolor{gray!20} LLaVA-Next 7B           & 64.9 & 1.9          & 70.2 & 2.9          & 64.2 & 2.1          \\
+FA                      & 64.9 & 2.7 (↑42.0\%)& 70.2 & 3.9 (↑36.4\%)& 64.2 & 3.6 (↑65.9\%)\\
FitPrune 40\%           & 64.9 & 2.9 (↑52.1\%)& 70.1 & 4.1 (↑44.1\%)& 64.2 & 3.3 (↑55.1\%)\\
FitPrune 40\% + FA      & 64.9 & 2.9 (↑55.3\%)& 70.1 & 4.6 (↑62.2\%)& 64.2 & 3.8 (↑76.6\%)\\
FitPrune 60\%           & 64.2 & 3.7 (↑97.9\%)& 70.1 & 5.1 (↑78.0\%)& 64.0 & 4.5 (↑107.9\%)\\
FitPrune 60\% + FA      & 64.2 & 3.7 (↑98.9\%)& 70.1 & 5.2 (↑80.0\%)& 64.0 & 5.0 (↑131.3\%)\\ 
\bottomrule
\end{tabular}
\caption{
Comparison of Flash Attention (FA), FitPrune, and their combinations on TextVQA, SQA, and GQA datasets. Accuracy is reported alongside throughput measured in samples per second. 
}
\label{tab:fitprune_flash_attention}
\end{table*}

\begin{figure}[!h]
\centering
\includegraphics[width=1.0\linewidth]{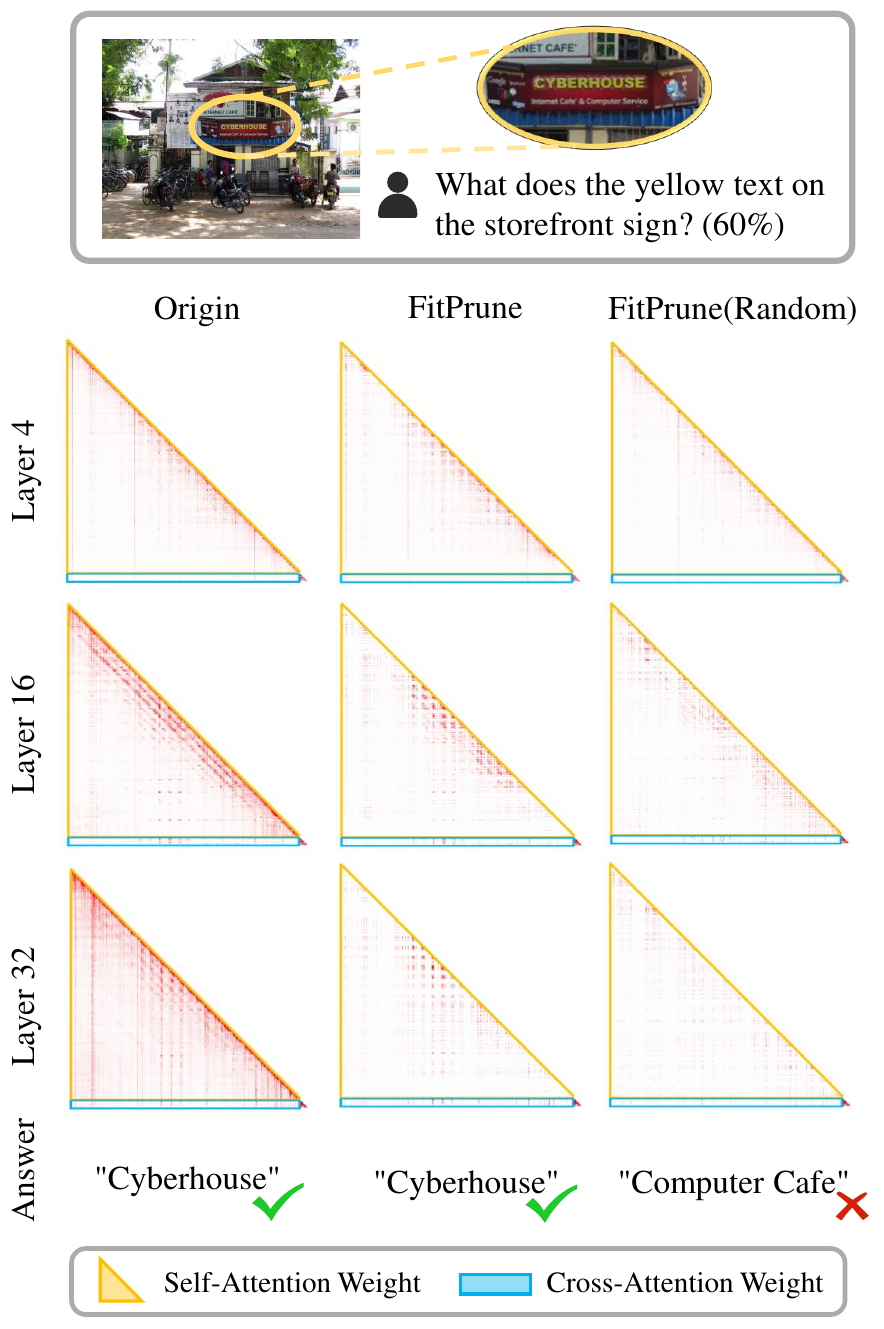}
\caption{Visualization of the attention map changes during the token removal process on the LLaVA-1.5 7B model. ``FitPrune(Random)'' indicates randomly pruning visual tokens according to the strategy generated by FitPrune. For comparison purposes, removed token regions are set to zero, and the system prompt part is omitted.}
\label{fig:atten_vis}
\vspace{-2mm}
\end{figure}

\subsection{Impact of pruning ratio on divergence and performance}
Fig.~\ref{fig:p_alpha} illustrates the relationship between pruning ratio and the average divergence $\alpha$ identified through our binary search process. As more tokens are pruned, $\alpha$ increases exponentially, indicating a growing divergence between distributions. Although this increase is more pronounced, the performance also declines correspondingly. This visualization highlights the crucial balance between reducing computational overhead and maintaining model accuracy, as discussed in our earlier analysis.

\begin{figure}[!t]
\centering
\includegraphics[width=1.0\columnwidth]{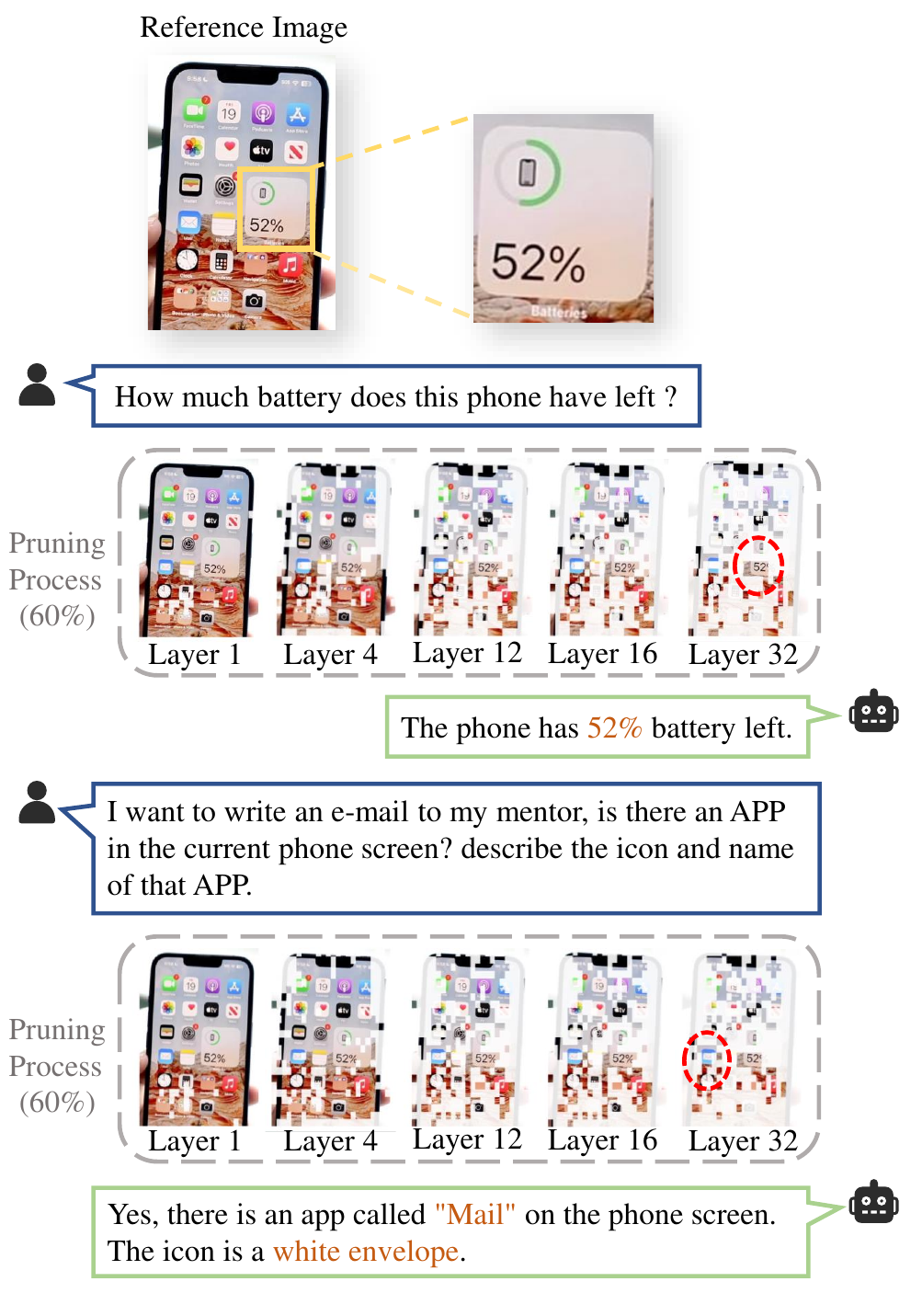}
\caption{
Example of LLaVA-1.5 7B using FitPrune at a 60\% pruning ratio on a mobile screenshot image. Despite the significant token reduction, the model's performance remains robust, highlighting the potential of FitPrune for real-world applications.
}
\label{fig:appendix_example1}
\vspace{-2mm}
\end{figure}

\begin{figure}[!t]
\centering
\includegraphics[width=1.0\columnwidth]{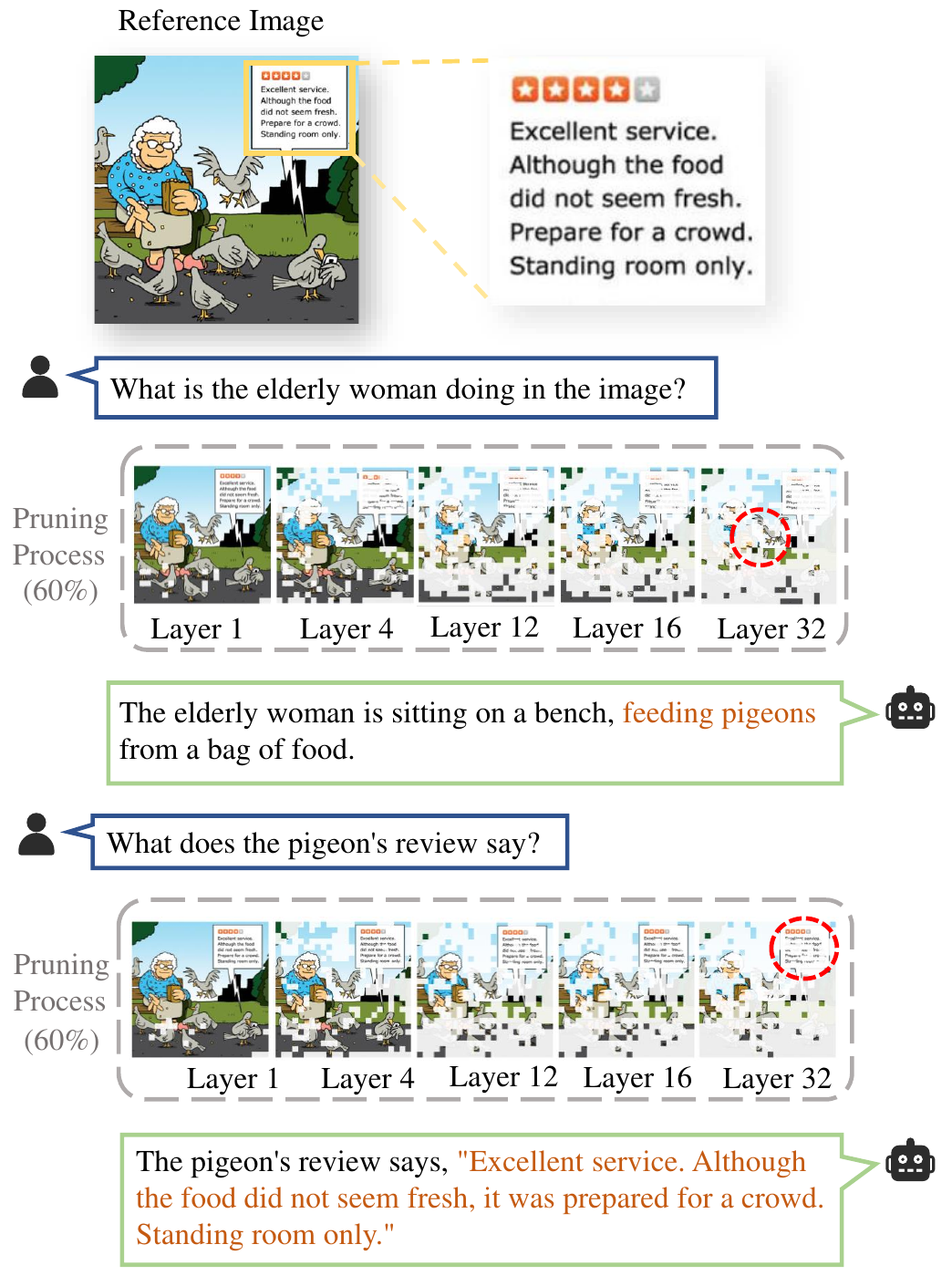}
\caption{
Example of LLaVA-1.5 7B using FitPrune at a 60\% pruning ratio on a comic-style image. The model's response highlights its ability to accurately interpret and understand both the visual details and textual elements. Notably, its OCR capability remains robust despite the substantial token reduction, underscoring the efficacy of FitPrune.
}
\label{fig:appendix_example2}
\end{figure}

\subsection{Detailed ablation results of pruning ratio}
In Tab.~\ref{tab:ablation_results} we report the results of different pruning ratio on the GQA benchmark. We can observe that our FitPrune method consistently maintains competitive performance even with extreme reductions in computational cost. For instance, at a 50\% pruning ratio, the model's accuracy only decreases by 0.1\%, with throughput increasing by 22.2\%. In extreme scenarios, our method continues to show robust performance. Specifically, at an 80\% pruning ratio, the model experiences a performance drop of 3.6\%, achieving a throughput increase of 40.2\% and reducing TFLOPs by 68.7\%. In summary, FitPrune demonstrates stability and practicality across varying reduction scenarios.

\subsection{Analysis of attention map changes} From Fig.~\ref{fig:atten_vis}, we can observe that after applying our method, approximately 25\% of the tokens are retained. The attention is concentrated in several continuous regions, which helps preserve crucial context and detail. In contrast, the attention map resulting from random token removal is more sparsely distributed and uniform, leading to the loss of critical image information and causing recognition errors.

\subsection{Ablation of attention distribution} From Tab.~\ref{tab:ablation_attention_distributions}, we observe that cross-attention is important for estimating visual tokens but is insufficient when used alone. Visual self-attention, despite being less effective individually, complements cross-attention when combined, achieving notable performance improvements across multiple benchmarks. This result validates the motivation and design of our FitPrune, highlighting the necessity of leveraging both attention mechanisms to better preserve critical information and enhance pruning effectiveness.  

\subsection{Ablation of FitPrune with Flash Attention} 
We conduct experiments to evaluate the acceleration effect of our FitPrune method, Flash Attention (FA) \cite{dao2022flashattention}, and their combinations across three benchmarks: TextVQA, SQA, and GQA. From the Tab.~\ref{tab:fitprune_flash_attention}, it can be observed that FitPrune achieves significant throughput improvements while maintaining accuracy. For instance, with a pruning ratio of 40\%, FitPrune increases throughput by 52.1\% on TextVQA and 78.0\% on SQA compared to the baseline, while preserving almost identical accuracy. When FitPrune is combined with FA, the speedup becomes even more remarkable. Specifically, the combination achieves a 131.3\% throughput gain on GQA with a pruning ratio of 60\%, significantly outperforming either method alone. These results demonstrate that FitPrune not only provides efficient acceleration on its own but also integrates seamlessly with FA to deliver superior performance.

\subsection{Generalization across different models} 
To evaluate the generalization ability of FitPrune, we test it on various MLLMs with different scales and families, including MobileVLMv2\cite{chu2024mobilevlm}, LLaVA-1.5\cite{liu2023visual}, QwenVL2\cite{wang2024qwen2}, and InternVL2\cite{chen2024internvl}. The results, shown in Tab.~\ref{tab:fitprune_generalization}, demonstrate the consistent effectiveness of FitPrune across these models. It can be observed that FitPrune achieves significant FLOPs reduction (pruning ratio of 60\%) while maintaining comparable performance. For example, on InternVL2 (8B), FitPrune achieves only a minor decrease in MMBench accuracy (82.4 to 81.8) while reducing FLOPs by 50.4\%. These results confirm that FitPrune generalizes well across both smaller models (e.g., MobileVLMv2) and larger models (e.g., LLaVA-1.5 13B), as well as different model families.

\vspace{2mm}
\subsection{Examples of multiturn dialogues}

Fig.~\ref{fig:appendix_example1} and Fig.~\ref{fig:appendix_example2} illustrate examples of multiturn dialogues using the LLaVA-1.5 7B model with our FitPrune method. Even with a high pruning ratio, the model maintains excellent performance, particularly in tasks requiring high visual detail, such as OCR. The visualizations of the pruning process show that the retained tokens are highly relevant to the questions, indicating effective information preservation. Notably, the model demonstrates robustness in handling complex visual reasoning, showcasing minimal performance degradation despite the reduction in computational resources. This suggests our pruning strategy preserves essential information while enhancing efficiency. Overall, it effectively balances performance and efficiency, proving its potential for complex multimodal tasks.

\end{document}